\documentclass[sigconf]{aamas} 
\usepackage{hyphenat}
\usepackage[dvipsnames]{xcolor}
\usepackage{placeins} 
\definecolor{MyDarkGreen}{rgb}{0.0, 0.4, 0.0}    
\definecolor{MyBrightGreen}{rgb}{0.0, 1.0, 0.0}  
\definecolor{MyCyanGreen}{HTML}{00A0A0}          

\usepackage{balance} 
\usepackage{pifont} 
\usepackage{algorithm} 
\usepackage{algorithmic} 
\usepackage{mdframed} 
\newcommand{\cmark}{\ding{51}} 

\setcopyright{ifaamas}
\acmConference[AAMAS '26]{Proc.\@ of the 25th International Conference
on Autonomous Agents and Multiagent Systems (AAMAS 2026)}{May 25 -- 29, 2026}
{Paphos, Cyprus}{C.~Amato, L.~Dennis, V.~Mascardi, J.~Thangarajah (eds.)}
\copyrightyear{2026}
\acmYear{2026}
\acmDOI{}
\acmPrice{}
\acmISBN{}

\acmSubmissionID{<<submission id>>}

\title{Structured Agent Distillation for Large Language Model Agents}
\author{Jun Liu\textsuperscript{1,4} \quad Zhenglun Kong\textsuperscript{2} \quad
Peiyan Dong\textsuperscript{3} \quad Changdi Yang\textsuperscript{4} \quad Tianqin Li\textsuperscript{1} \quad Yanyue Xie\textsuperscript{4} \quad Yifan Gong\textsuperscript{5} \quad Xuan Shen\textsuperscript{4} \quad Pu Zhao\textsuperscript{4} \quad Hao Tang\textsuperscript{1,6} \quad Geng Yuan\textsuperscript{7} \quad Wei Niu\textsuperscript{7} \\ Wenbin Zhang\textsuperscript{8} \quad Xue Lin\textsuperscript{4} \quad Yanzhi Wang\textsuperscript{4} \quad Dong Huang\textsuperscript{1} \\
\textsuperscript{1}Carnegie Mellon University, USA\quad
\textsuperscript{2}Harvard University, USA\quad
\textsuperscript{3}MIT, USA\quad\\
\textsuperscript{4}Northeastern University, USA\quad
\textsuperscript{5}Adobe Research, USA\quad
\textsuperscript{6}National University of Singapore, Singapore\quad\
\textsuperscript{7}University of Georgia, USA\quad
\textsuperscript{8}Florida International University, USA\\
}

\begin{abstract}
Large language models (LLMs) exhibit strong capabilities as decision-making agents by interleaving reasoning and actions, as seen in ReAct-style frameworks.
Yet, their practical deployment is constrained by high inference costs and large model sizes.
We propose \textbf{Structured Agent Distillation}, the first framework to distill a ReAct-based LLM agent into a smaller model while preserving both reasoning fidelity and action consistency. Our method introduces a structured, span-level distillation strategy that explicitly segments trajectories into reasoning and action spans, enabling fine-grained alignment beyond standard token-level imitation.
Unlike other advanced distillation methods, 
Our method segments trajectories into \texttt{[REASON]} and \texttt{[ACT]} spans, applying segment-specific losses to align each component with the teacher's behavior.
This structure-aware supervision enables compact agents to better replicate the teacher’s decision process.
Experiments on ALFWorld, HotPotQA-ReAct, and WebShop show that our approach consistently outperforms token-level and imitation learning baselines, achieving significant compression with minimal performance drop.
Scaling and ablation results further highlight the importance of span-level alignment for efficient and deployable agents. We will release code upon acceptance.
\end{abstract}

\keywords{Agent Distillation, Span-Level Alignment, Reasoning and Action Segmentation, Large Language Models (LLMs), Span-Level Alignment}

\newcommand{\BibTeX}{\rm B\kern-.05em{\sc i\kern-.025em b}\kern-.08em\TeX}

\begin{document}


\pagestyle{fancy}
\fancyhead{}


\maketitle 


\section{Introduction}

Large language models (LLMs) have recently been extended beyond language modeling into decision-making roles, giving rise to \emph{LLM-based general agents}—systems that solve complex tasks by interleaving multi-step reasoning and tool-augmented actions. As these agents move toward real-world deployment, platform designers face inherent tradeoffs between efficiency and broader system objectives~\cite{athey2022smiles,ma2025balancing}. Frameworks like ReAct~\cite{yao2023react}, Toolformer~\cite{schick2023toolformer}, and WebGPT~\cite{nakano2021webgpt} demonstrate that LLMs can operate through structured \emph{reasoning-action trajectories}—sequences alternating between deliberation and execution to complete tasks such as planning, web navigation, and multi-hop question answering.
Chain-of-thought (CoT) prompting~\cite{wei2022chain,wang2022self} encourages models to decompose complex tasks into intermediate reasoning steps before acting, reinforcing the need to preserve reasoning–action structure during training.

Despite their effectiveness, LLM-based general agents remain costly to deploy due to model size and inference overhead. To address this, recent work distills large agents into smaller student models. However, most approaches rely on \emph{token-level supervision}~\cite{hinton2015distilling, turc2019well, eldan2023tinystories, gu2024minillm,minixhofer2025cross,cui2025multi,zhang2024knowledge}, which treats the agent trajectory as a flat token sequence and aligns predictions step by step—ignoring its structured composition of reasoning and action.

\noindent \textbf{Limitations of Token-Level Distillation.}
This paradigm fails to capture the \textit{structural nature} of agent behavior:
(i) it overlooks long-range dependencies between reasoning and action~\cite{zheng2024sglang};
(ii) it lacks span-level supervision, blurring the distinction between planning and execution;
(iii) it causes semantic drift during rollouts, degrading coherence and task success.

\noindent \textbf{Our Approach: Structured Agent Distillation.}
We propose a structure-aware compression framework that explicitly models the compositional structure of agent behavior.
Our method segments each trajectory into \texttt{[REASON]} and \texttt{[ACT]} spans and supervises them with span-specific objectives.
By applying segment-aware masking and reasoning–action alignment, our approach preserves both the rationale and the resulting decision—enabling more faithful and coherent student agents.

\noindent\textbf{Our Contributions:}
\begin{itemize}
    \item Our work is the first to distill ReAct-style LLM agents using structured span-level supervision: we segment trajectories into reasoning and action spans and apply span-specific alignment via token-level masks, improving over naive token-level distillation.
    \item We validate our approach on ALFWorld, HotPotQA-ReAct, and WebShop, achieving consistent gains in task success, planning efficiency, and chain-of-thought (CoT) alignment over strong token-level baselines.
    \item We conduct comprehensive scaling and ablation studies, demonstrating that segment-level supervision is critical for training compact and robust student agents.
\end{itemize}


\section{Motivation}

\subsection{Why Token-Level Distillation Falls Short}
\textit{LLMs vs. General Agents.}
While LLMs focus on single-turn generation, LLM-based general agents operate in interactive settings where structured reasoning and action unfold over multiple steps. Token efficiency in agents must therefore consider trajectory-level~\cite{reedgeneralist,shinn2023reflexion} latency and semantic role differentiation. Table~\ref{tab:llm_vs_agent} summarizes the key distinctions between the two paradigms.

This structured nature of agent trajectories poses unique challenges for compression and acceleration.
In particular, existing methods such as token-level distillation~\cite{hinton2015distilling}, originally designed for next-token prediction, fail to capture the hierarchical nature of agent behavior.

Token-level distillation supervises the student at each decoding step using cross-entropy~\cite{cover1999elements} or KL divergence~\cite{kullback1951information} between teacher and student outputs. While this is effective for language modeling, it fails to account for the structured nature of agent trajectories—specifically the distinction between intermediate reasoning and final action execution.

Critically, \emph{token-level methods lack structural awareness}, treating all tokens equally without distinguishing their functional roles in the agent trajectory. 
In practice, trajectories often alternate between internal reasoning steps and external actions—two semantically distinct spans that require different forms of supervision.

As a result, the student learns to match surface-level actions while ignoring the underlying rationale, often skipping key planning steps required to complete the task.

\begin{table*}[ht] \small
\centering
\caption{Comparison between LLMs and LLM-based General Agents in terms of token efficiency}
\label{tab:llm_vs_agent}
\begin{tabular}{lccc}
\toprule
\textbf{Dimension} & \textbf{LLMs} & \textbf{LLM-based General Agents} \\
\midrule
Application & Static, single-turn tasks  (QA, etc.) &Multi-turn interactive tasks (e.g., WebNav) \\
Objective & Minimize FLOPs in generation & Mini token budget across reasoning + action \\
Token Source & Fixed input tokens per prompt & Dynamic tokens in reasoning-action steps\\
Latency Target & Single-step inference efficiency &End-to-end efficiency over agent steps \\
Output Format & Final text sequence &Task trajectory  (reasoning steps + actions) \\

\bottomrule
\end{tabular}
\end{table*}

\subsection{Motivation: Toward \textbf{Structured Agent Distillation}}

We propose \textbf{Structured Agent Distillation (SAD)}, which segments trajectories into \texttt{[REASON]} and \texttt{[ACT]} spans and applies span-specific supervision to improve structural imitation.
A curriculum mechanism further enhances stability by ordering training examples by complexity.

Table~\ref{tab:framework_comparison} summarizes representative LLM-based agent frameworks in terms of four dimensions: external tool usage, reasoning-action alignment, segment-aware supervision, and curriculum-guided training. While prior methods such as ReAct and Voyager support structured reasoning and tool use, they lack segment-level supervision and curriculum scheduling. In contrast, our \textbf{Structured Agent Distillation} framework uniquely supports all four dimensions, enabling more faithful and efficient agent compression.

\begin{table}[t]
\small
\centering
\caption{
Comparison of LLM agent training frameworks. Only our method supports all four dimensions of structured agent distillation.
\medskip
\textbf{Tool}: supports external API calls or tool use;
\textbf{R--A Align.}: aligns structured reasoning and action spans;
\textbf{Seg.-aware Sup.}: applies supervision across reasoning–action sequences;
\textbf{Curric.}: uses trajectory difficulty for progressive training~\cite{kumar2010self}.
}
\label{tab:framework_comparison}
\renewcommand{\arraystretch}{1.15}
\setlength{\tabcolsep}{3.5pt}
\begin{tabular}{lcccc}
\toprule
\textbf{Framework} & \textbf{Tool} & \textbf{R--A Align.} & \textbf{Seg.-aware Sup.} & \textbf{Curric.} \\
\midrule
Token-Level KD~\cite{hinton2015distilling} & \textcolor{red}{\ding{55}} & \textcolor{red}{\ding{55}} & \textcolor{red}{\ding{55}} & \textcolor{red}{\ding{55}} \\
ReAct~\cite{yao2023react} & \cmark & \cmark & \textcolor{red}{\ding{55}} & \textcolor{red}{\ding{55}} \\
Toolformer~\cite{schick2023toolformer} & \cmark & \textcolor{red}{\ding{55}} & \textcolor{red}{\ding{55}} & \textcolor{red}{\ding{55}} \\
Voyager~\cite{wang2023voyager} & \cmark & \cmark & \textcolor{red}{\ding{55}} & \textcolor{red}{\ding{55}} \\
\textbf{SAD (Ours)} & \cmark & \cmark & \cmark & \cmark \\
\bottomrule
\end{tabular}
\vspace{-10pt}
\end{table}

\section{Structured Agent Distillation Framework}
\begin{figure}[t]
  \centering
  \includegraphics[width=0.60\linewidth]{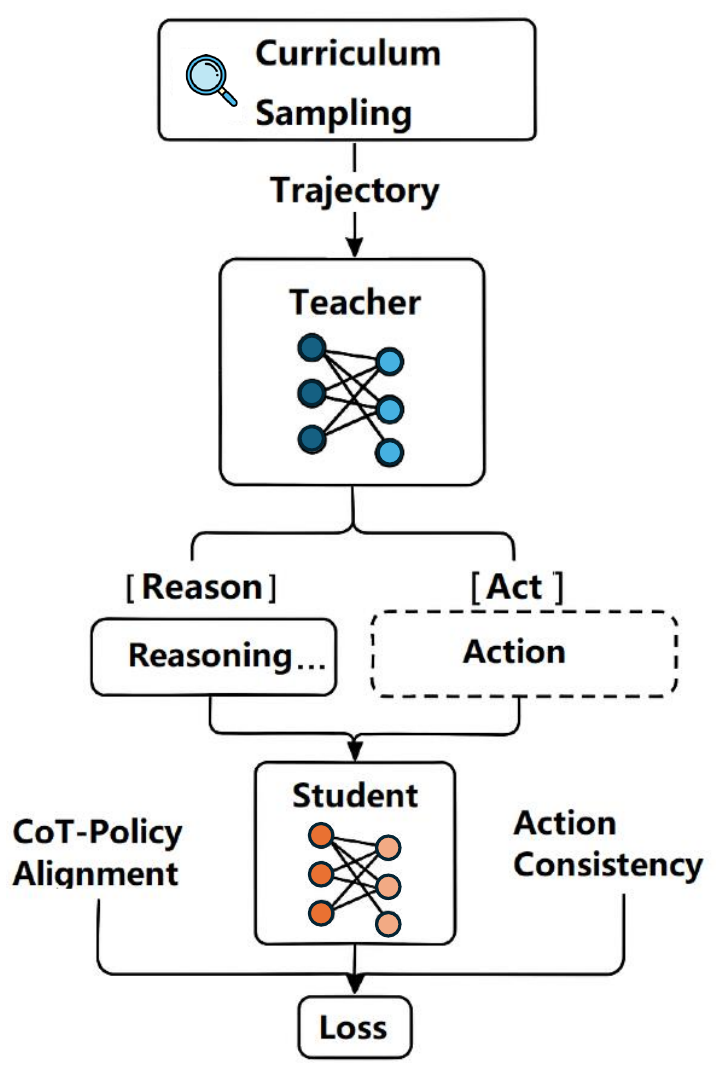}
  \caption{Structured Agent Distillation framework. The teacher provides segmented reasoning–action trajectories. The student aligns CoT traces and actions via span-specific KL losses, with projected gradients and curriculum sampling for stability.
  }
  \Description{Structured Agent Distillation framework.}
  \label{fig:framework}
  \vspace{-15pt}
\end{figure}

\textbf{Structured Agent Distillation} (SAD) segments teacher trajectories into reasoning (\texttt{Reason}) and interaction (\texttt{Action/Observation}) spans, each supervised independently to promote phase-specific alignment.



As shown in Figure~\ref{fig:framework}, the teacher agent, given an observation and task prompt, produces \texttt{[REASON]}ing traces and \texttt{[ACT]}ion outputs, forming a trajectory $\tau = (\texttt{reason}, \texttt{action})$ used for curriculum sampling.

The student learns from $\tau$ via two losses: 
(1) \emph{CoT-Policy Alignment} $\mathcal{L}_{\text{CoT}}$ aligns reasoning, and 
(2) \emph{Action Consistency} $\mathcal{L}_{\text{Act}}$ aligns decisions.

Refer to Appendix~\ref{appendix:framework} for comprehensive analysis.

\subsection{Problem Formulation}
\label{sec:problemformulation}
We aim to distill high-capacity ReAct-style teacher agents into smaller student models while preserving structured decision-making behavior.
Each teacher's trajectory is a sequence of interleaved reasoning and action components:
\begin{equation}
\tau = \left[ (r_1, r_2, \ldots, r_k), (a_1, a_2, \ldots, a_m) \right],
\label{eq:tau_structure}
\end{equation}
where $r_i \in \mathcal{R}$ are reasoning tokens (e.g., CoT steps), and $a_j \in \mathcal{A}$ are action tokens (e.g., tool calls, answers).

Given a teacher policy $\pi_T(\tau)$, the goal is to train a compact student policy $\pi_\theta(\tau)$ such that
\begin{equation}
\pi_\theta(\tau) \approx \pi_T(\tau),
\end{equation}
preserving both semantic reasoning and execution structure beyond token-level matching.

To enable sequence-to-sequence modeling, we linearize each trajectory into a flattened form with segment markers:
\[
\tau' = \texttt{[REASON]} \ r_1 \cdots r_k \ \texttt{[ACT]} \ a_1 \cdots a_m.
\]
We tokenize this as 
\begin{equation}
x = \texttt{Tokenize}(\tau') = (x_1, x_2, \ldots, x_T),
\label{eq:tokenx}
\end{equation} and assign each token $x_t$ a segment label $s_t \in \{\text{Reason}, \text{Action}\}$, indicating its span. These labels are used to compute segment-aware losses during training.

For clarity, we adopt the following notation:
$\tau$ denotes the structured reasoning–action trajectory,
$\tau'$ its linearized form with explicit segment markers,
and $x=\texttt{Tokenize}(\tau')$ the token sequence processed by the model.
Accordingly, $\pi_\theta$ always operates on tokenized inputs $x$,
while $\tau$ and $\tau'$ are used only for segmentation and mask construction.

\subsection{Trajectory Segmentation}
\label{sec:Trajectory}
Given a teacher-generated trajectory $\tau$, we decompose it into two disjoint spans:

\[
(\tau^{(r)}, \tau^{(a)}) \leftarrow \text{Segment}(\tau),
\]

where $\tau^{(r)}$ denotes the reasoning span and $\tau^{(a)}$ denotes the action span.
This segmentation is performed via lightweight rule-based parsing based on prompt templates consistent across tasks (see Appendix~D).
The segmented trajectory is then tokenized into a sequence $x$ defined in Eq. (\ref{eq:tokenx}). 

While the above formulation assumes a single reasoning–action pair,
SAD naturally extends to multi-step ReAct trajectories 
$\tau = [(r^{(i)}, a^{(i)}, o^{(i)})]_{i=1}^{K}$ 
by applying union masks over multiple reasoning/action spans.
A detailed construction and example are provided in Appendix~\ref{app:multistep}.

\subsection{Structured Agent Distillation Objectives}
\label{sec:structured_objectives}

To supervise student agents under \textbf{SAD}, we align binary token masks $m_r(t)$ and $m_a(t)$ with the tokenized sequence $x = (x_1, \ldots, x_T)$, enabling selective supervision over structurally distinct parts of the trajectory.


\paragraph{(1) CoT-Policy Alignment Loss.}
For reasoning tokens, the student’s conditional distribution 
$p_S(\cdot \mid x_{<t})$ is aligned with the teacher’s distribution 
$p_T(\cdot \mid x_{<t})$ using Kullback–Leibler(KL) divergence:
\begin{equation}
\mathcal{L}_{\text{CoT}} = \sum_{t=1}^{T} m_r(t)\, \mathrm{KL}\!\left(p_T(\cdot \mid x_{<t}) \,\|\, p_S(\cdot \mid x_{<t})\right)
\label{eq:colt}
\end{equation}

\paragraph{(2) Action Consistency Loss.}
For action tokens, we similarly minimize KL divergence:
\begin{equation}
\mathcal{L}_{\text{Act}} = \sum_{t=1}^{T} m_a(t)\, \mathrm{KL}\!\left(p_T(\cdot \mid x_{<t}) \,\|\, p_S(\cdot \mid x_{<t})
\right)
\label{eq:act}
\end{equation}

CoT-Policy Alignment operates over the full vocabulary during \texttt{[REASON]} spans to guide the student’s intermediate reasoning steps, encouraging alignment with the teacher’s chain-of-thought. In contrast, Action Consistency Loss applies KL over a discrete action space during \texttt{[ACT]} spans, ensuring that the student replicates the teacher’s action decisions. 

\paragraph{Final Objective.}
The total structured loss aggregates these terms:
\begin{equation}
\mathcal{L}_{\text{total}} = \lambda_r \cdot \mathcal{L}_{\text{CoT}} + \lambda_a \cdot \mathcal{L}_{\text{Act}},
\label{eq:total}
\end{equation}
where $\lambda_r$ and $\lambda_a$ are scalar weights balancing the two objectives. We set $\lambda_r = \lambda_a = 1.0$ to equally weight reasoning and action supervision in the final loss.

\noindent\textbf{Clarification.}  
Although the loss terms are unscaled, this formulation is \emph{not} equivalent to computing a single 
token-level KL over the entire vocabulary.  
Token-level KL normalizes over the joint token space $\mathcal{V}_r \cup \mathcal{V}_a$, 
which couples gradients from frequent reasoning tokens and rare but critical action tokens.  
In contrast, SAD applies KL over disjoint normalization domains 
($\mathcal{V}_r$ for reasoning, $\mathcal{V}_a$ for action), 
thereby changing both the normalization space and gradient direction.  
This decomposition alters the optimization geometry and prevents cross-span interference, 
constituting a fundamental difference from a flat token-level KL even when $\lambda_r$ and $\lambda_a$ are equal.
While our formulation assumes teacher-forced alignment during training, the loss can be extended to accommodate mismatched trajectories via alignment-based matching, as detailed in Appendix ~\ref{app:alignment}.


\subsection{Optimization View: Gradient Projection and Span Decoupling}
\label{sec:insight}

\begin{figure}[t]
\centering
\includegraphics[width=0.38\textwidth]{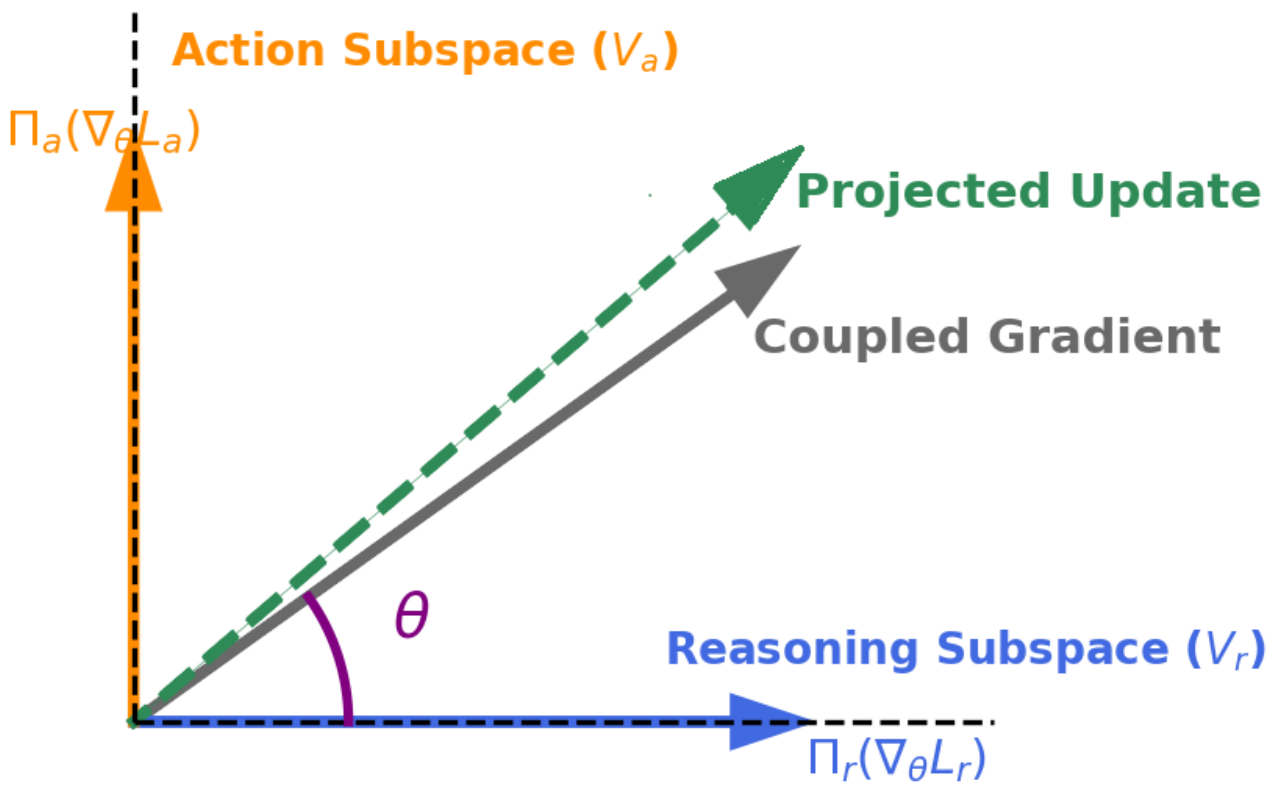}
\vspace{-10pt}
\caption{
\textbf{Optimization structure of Structured Agent Distillation (SAD) via gradient projection.}
Standard token-level KL (gray) couples reasoning and action gradients, leading to a conflict angle $\theta$.
SAD resolves this by projecting gradients onto reasoning ($\mathcal{V}_r$) and action ($\mathcal{V}_a$) subspaces,
forming an orthogonal decomposition.
The resulting projected update (green) follows the composite direction 
$\Pi_{\text{reason}}(\nabla_\theta \mathcal{L}_{\text{CoT}})
+\Pi_{\text{action}}(\nabla_\theta \mathcal{L}_{\mathrm{Act}})$, 
eliminating cross-span interference and clarifying SAD’s optimization geometry. }
\Description{Diagram showing geometric interpretation of Structured Agent Distillation (SAD), illustrating gradient projections and subspaces.}

\label{fig:grad_projection}
\vspace{-16pt}
\end{figure}

To better understand the functional benefit of SAD, we present an \emph{optimization-based interpretation} of how span-specific KL losses reshape the gradient landscape.  
Rather than assuming a cognitive or semantic split, SAD introduces an optimization structure that avoids interference between 
reasoning and action supervision signals.

\textbf{As illustrated in Figure~\ref{fig:grad_projection}}, 
a token-level KL couples heterogeneous gradients into a single update direction (gray arrow),
creating a conflict angle~$\theta$ between reasoning and action forces in parameter space.  
SAD reformulates this process as an orthogonal gradient projection: 
reasoning and action components are separately normalized in their respective subspaces 
($\mathcal{V}_r$, $\mathcal{V}_a$) and then recombined geometrically.  
This projection removes cross-span interference and yields span-specific updates, 
altering the overall optimization geometry.

Let $\mathcal{V}_r$ and $\mathcal{V}_a$ denote the token domains for reasoning and action,
along with $\mathcal{V}_r \cap \mathcal{V}_a = \varnothing$.
A standard token-level KL minimizes
\begin{equation}
\nabla_\theta \mathcal{L}_{\text{token}}
= \nabla_\theta\,\mathrm{KL}\!\left(p_T(\cdot\mid x_{<t}) \,\|\, p_S(\cdot\mid x_{<t})\right),
\end{equation}
where normalization over the entire vocabulary $\mathcal{V}_r\cup\mathcal{V}_a$
forces shared probability mass between semantically incompatible tokens.
This coupling biases gradients toward frequent reasoning tokens
and suppresses rare but task-critical action tokens, 
explaining why a single distribution cannot disentangle these behaviors.

\textbf{SAD} resolves this by restricting the KL computation to disjoint subspaces:
\begin{equation}
\nabla_\theta \mathcal{L}_{\text{SAD}}
= m_r(t)\,\nabla_\theta \mathrm{KL}_{\mathcal{V}_r}\!\left(p_T \,\|\, p_S\right)
+ m_a(t)\,\nabla_\theta \mathrm{KL}_{\mathcal{V}_a}\!\left(p_T \,\|\, p_S\right),
\end{equation}
which performs \emph{gradient projection} onto reasoning and action subspaces,
\begin{equation}
\nabla_\theta \mathcal{L}_{\text{SAD}}
= \Pi_{\text{reason}}\!\big(\nabla_\theta  \mathcal{L}_{\text{CoT}}\big)
+ \Pi_{\text{action}}\!\big(\nabla_\theta \mathcal{L}_{\mathrm{Act}}\big).
\end{equation}
This projection changes both the normalization domain and gradient direction,
eliminating cross-span interference and yielding span-specific updates.
\textbf{Hence, the distinction from token-level KL is geometric, not cosmetic:}
SAD introduces a structure-aware gradient decomposition rather than simply applying
the KL divergence to a smaller subset of tokens.

\subsection{Multi-Step ReAct Trajectories and Multi-Span Masks}
\label{sec:multistep}

\noindent\textbf{Trajectory Model.}
Extending the single-step formulation in Eq.~(\ref{eq:tau_structure}),
we generalize SAD to multi-step ReAct episodes composed of alternating reasoning, action, and observation segments:
\begin{equation}
\tau = \big[(r^{(1)}, a^{(1)}, o^{(1)}), (r^{(2)}, a^{(2)}, o^{(2)}), \ldots, (r^{(K)}, a^{(K)}, o^{(K)})\big],
\end{equation}
where $r^{(i)}$, $a^{(i)}$, and $o^{(i)}$ denote the reasoning trace, executed action, and subsequent observation at step $i$.
Linearization yields
\[
\tau' = \prod_{i=1}^{K} \big(\texttt{[REASON]}~r^{(i)}~
\texttt{[ACT]}~a^{(i)}~
\texttt{[OBS]}~o^{(i)}\big), \qquad
x = \texttt{Tokenize}(\tau').
\]
This formulation extends SAD beyond the single-step assumption, enabling structured supervision across multiple reasoning–action cycles.

\noindent\textbf{Segment-Aware Mask Construction.}
Each token $x_t$ in the sequence is assigned a binary membership mask:
\begin{equation}
\begin{aligned}
m_r(t) &= \mathbf{1}[x_t \in \cup_i r^{(i)}], 
m_a(t) &= \mathbf{1}[x_t \in \cup_i a^{(i)}], \\
m_o(t) &= \mathbf{1}[x_t \in \cup_i o^{(i)}],
\end{aligned}
\end{equation}

We enforce an \emph{exactly-one} constraint for every token:
\begin{equation}
m_r(t) + m_a(t) + m_o(t) = 1, \quad \forall t \in [1, T],
\end{equation}
ensuring non-overlapping and functionally disjoint spans. 
Reasoning and action masks ($m_r, m_a$) are used for supervision, while observation masks ($m_o$) indicate environmental feedback.

\noindent\textbf{Observation Handling.}
Each token is assigned to \emph{exactly one} of the three functional categories—reasoning, action, or observation—ensuring disjoint span boundaries. 
However, only reasoning and action tokens contribute to the distillation loss. Observation tokens ($m_o(t)=1$) are excluded from the distillation loss, as they encode deterministic feedback from the environment rather than agent behavior. 
This exclusion prevents the student from overfitting to static textual observations and focuses learning on reasoning and decision quality.
Nonetheless, the framework permits optional extensions:
(1) adding an auxiliary cross-entropy term for perceptual grounding, or 
(2) defining a separate observation head for multimodal tasks. 
All reported experiments adopt the exclusion setting.

\noindent\textbf{Supervision over Multi-Span Masks.}
Structured losses defined in Section~\ref{sec:structured_objectives} 
(Eq.~(\ref{eq:colt}) and~(\ref{eq:act}))
are applied independently to each reasoning and action span, as detailed in Appendix~\ref{app:action_space}. with
$m_r$ and $m_a$ computed as the union of all \texttt{[REASON]} and \texttt{[ACT]} segments.
This union-mask construction generalizes SAD to multi-turn reasoning–acting trajectories,
preserving disjoint functional roles and preventing cross-span gradient interference.

\noindent\textbf{Illustrative Example.}
A two-step episode ($K{=}2$) from ALFWorld:
\begin{mdframed}[backgroundcolor=gray!5, linewidth=0.5pt, roundcorner=5pt]
\small
\texttt{[REASON]} I will first check the table. 
\texttt{[ACT]} search[tray] 
\texttt{[OBS]} You see a tray.\\
\texttt{[REASON]} Now I will pick it up. 
\texttt{[ACT]} pickup[tray] 
\texttt{[OBS]} Tray in inventory.
\end{mdframed}
Here, reasoning tokens correspond to $m_r{=}1$, action tokens to $m_a{=}1$, and observation tokens to $m_o{=}1$. 
Only reasoning and action tokens receive gradient updates, clarifying span-level supervision semantics and aligning with multi-turn ReAct agent behavior.

\subsection{Optimization Analysis and Intuition}
\label{sec:analysis}

\textbf{Why it works in practice.}
The decoupled objectives alter both gradient magnitude and direction across spans.
Empirically, SAD reduces step-to-step gradient variance and improves training stability under limited data.
This acts as an \emph{implicit curriculum}: reasoning spans provide dense signals for high-level coherence,
while action spans yield sparse but decisive grounding signals.
The balanced supervision accelerates convergence and strengthens reasoning–action compositionality.

SAD provides a \emph{principled change in optimization geometry} that
(1) moves beyond simplified cognitive assumptions,
(2) explains why a single token-level distribution fails to separate reasoning and action,
and (3) establishes a structure-aware, theoretically grounded distillation objective.

\subsection{Semantic Decoupling and Example}

While the overall loss is additive in form, our supervision is fundamentally different from flat token-level imitation. We explicitly decompose the learning signal into structurally disjoint spans—\texttt{[REASON]} and \texttt{[ACT]}—and apply segment-specific losses to each, preserving the semantics of multi-phase agent behavior.

\noindent\textbf{CoT-Policy Alignment Loss} ($\mathcal{L}_{\text{CoT}}$) supervises predictions within the reasoning span, promoting coherent multi-step inference aligned with the teacher's thought patterns. \textbf{Action Consistency Loss} ($\mathcal{L}_{\text{Act}}$) applies only to the action span, enforcing the accurate replication of grounded decisions.

Each token is assigned to exactly one functional span using binary masks $\{m_r(t), m_a(t)\}$, which gate gradient flow. This masking enforces \emph{semantic separation} during training, ensuring that the student independently learns high-level reasoning and low-level execution. Unlike token-level KL with soft targets, our structure-aware formulation avoids loss interference across phases, better modeling causal dependencies (reason $\rightarrow$ act). Ablations confirm that removing either component harms performance.

\noindent\textbf{Example.}

\begin{mdframed}[backgroundcolor=gray!5, linewidth=0.5pt, roundcorner=5pt]
\small
\textbf{Instruction:}  \texttt{"Find the tray"} \\
\textbf{Teacher:} \texttt{[REASON]} \texttt{"I will first look on the table to check for a tray..."} $\rightarrow$ \texttt{[ACT]} \texttt{search[tray]} \\
\textbf{Student:}  \texttt{[REASON]} \texttt{"Maybe it's on the shelf, I should check there."} $\rightarrow$ \texttt{[ACT]} \texttt{search[tray]}
\end{mdframed} 

Although the student executes the correct action, their reasoning deviates from the teacher’s thought process. $\mathcal{L}_{\text{CoT}}$ penalizes semantic deviations within \texttt{[REASON]} via KL divergence between teacher and student token distributions, producing gradients that align multi-step reasoning. $\mathcal{L}_{\text{Act}}$ rewards correct predictions in \texttt{[ACT]}, allowing action alignment even when reasoning differs.

\subsection{Curriculum Sampling in Structured Distillation}

To further enhance learning efficiency and stability, we employ curriculum learning~\cite{bengio2009curriculum,guo2018curriculumnet} based on a trajectory complexity score:
\begin{equation}
C(\tau) = \alpha \cdot \mathrm{len}(r_{1:k}) + \beta \cdot \mathrm{len}(a_{1:m}) + \gamma \cdot \mathrm{entropy}\!\left(\pi_T(\tau)\right),
\end{equation}
where $\mathrm{len}(r_{1:k})$ and $\mathrm{len}(a_{1:m})$ denote the lengths of the reasoning and action segments, respectively, and $\mathrm{entropy}(\pi_T(\tau))$ reflects teacher uncertainty. The weights $\alpha, \beta, \gamma$ balance their relative contributions. During training, trajectories are sorted by $C(\tau)$, allowing the model to start with examples and gradually progress to more complex ones. Detailed analysis in Appendix~\ref{appendix:curriculum}.

\subsection{Training Algorithm}

Algorithm~\ref{alg:structured_distill} outlines our structured agent distillation process. 
The student policy $\pi_\theta$ learns to imitate the teacher $\pi_T$ across reasoning stages, guided by a curriculum scheduler $\mathcal{C}$ that samples increasingly complex trajectories. 
Each trajectory is tokenized into reasoning and action spans, and the student predicts tokens autoregressively. 
The objective aggregates reasoning and action losses, and gradients are backpropagated to update $\theta$.

\begin{algorithm}[t]
\caption{Structured Agent Distillation (SAD)}
\label{alg:structured_distill}
\begin{algorithmic}[1]
\STATE Initialize teacher policy $\pi_T$, student policy $\pi_\theta$, and curriculum scheduler $\mathcal{C}$
\FOR{epoch $= 1$ to $E$}
    \STATE Sample multi-step trajectory $\tau = \{(r^{(i)}, a^{(i)}, o^{(i)})\}_{i=1}^K \sim \mathcal{C}$
    \STATE Linearize and tokenize $\tau' = \texttt{[REASON]}~r^{(i)}~\texttt{[ACT]}~a^{(i)}~\texttt{[OBS]}~o^{(i)}$
    \STATE Construct binary masks $m_r, m_a, m_o$ over tokens $x_{1:T}$
    \STATE Forward pass: obtain student logits $p_S(\cdot\mid x_{<t})$
    \STATE Compute losses $\mathcal{L}_{\text{CoT}}, \mathcal{L}_{\text{Act}}$
    \STATE Aggregate weighted total loss 
        $\mathcal{L}_{\text{total}} = \lambda_r \mathcal{L}_{\text{CoT}} + \lambda_a \mathcal{L}_{\text{Act}}$
    \STATE Update parameters 
        $\theta \leftarrow \theta - \eta\,\nabla_\theta \mathcal{L}_{\text{total}}$
\ENDFOR
\end{algorithmic}
\end{algorithm}

\section{Experiments}

\subsection{Experimental Setups}
\label{sec:setup}

\noindent\textbf{Agent Environments.}
We evaluate on three representative benchmarks:
\textbf{~\ding{172} } Embodied benchmark ALFWorld~\cite{ALFWorld20} for embodied instruction following,
\textbf{~\ding{173} } Web benchmark WebShop~\cite{yao2022webshop} towards scalable real-world web interaction with grounded language agents, and  
\textbf{~\ding{174} } multi-hop question benchmarks HotPotQA-ReAct~\cite{yao2023react} for multi-hop QA with reasoning traces. ALFWorld and WebShop test structured decision-making, while HotPotQA-ReAct 
requires open-ended multi-hop reasoning and free-form answer generation—thus 
already covering both discrete and natural-language modalities. 
All reported results are averaged over 5 independent runs.

\noindent\textbf{Baselines.} The baselines include token level KD~\cite{gu2024minillm}, word-level \textbf{KD}~\cite{distilbert,lightpaff}, and \textbf{SeqKD}~\citep{skd,alpaca,ITGPT4,lima}.

\noindent\textbf{Dataset Statistics.}
We adopt standard splits from existing ReAct-based benchmarks.
ALFWorld comprises 8{,}055 instruction-following trajectories 
(5{,}400 train / 1{,}200 val / 1{,}475 test).  
WebShop contains 12{,}000 web interaction samples 
(8{,}000 / 2{,}000 / 2{,}000),  
and HotPotQA-ReAct includes 90{,}447 multi-hop reasoning examples 
(84{,}000 / 3{,}447 / 3{,}000).

The details of Experimental Setups in Appendix~\ref{appendix:task-setup}.

\subsection{Experimental Results and Analysis}

We evaluate student agent performance across three benchmarks—
ALFWorld, WebShop,
and HotPotQA-ReAct—comparing our proposed \textbf{Structured Agent Distillation} with the baseline.\\
We report results across three evaluation metrics: task success rate, reasoning efficiency, and CoT consistency (Table~\ref{tab:all_metrics_combined}).

\begin{table*}[t]  \small
\centering
\caption{
Unified comparison of task success (↑), reasoning length (↓), CoT match (↑), and episode latency (steps) (↓) across ALFWorld, WebShop, and HotPotQA. The teacher is a GPT-2-1.5B ReAct-style agent. Students trained via \textbf{Structured Agent Distillation} consistently outperform token-level KD~\cite{gu2024minillm}, KD~\cite{distilbert,lightpaff}, and SeqKD~\cite{skd,vicuna,alpaca,ITGPT4,lima} baselines.
} %
\vspace{-5pt}
\label{tab:all_metrics_combined}
\resizebox{0.8\textwidth}{!}{
\begin{tabular}{l|ccc|ccc|ccc|ccc}
\toprule
\textbf{Method} & \multicolumn{3}{c|}{\textbf{Task Success ↑}} & \multicolumn{3}{c|}{\textbf{Reasoning Length ↓}} & \multicolumn{3}{c|}{\textbf{CoT Match Rate ↑}} & \multicolumn{3}{c}{\textbf{Episode Latency ↓}} \\
& ALF & Web & Hot & ALF & Web & Hot & ALF & Web & Hot & ALF & Web & Hot \\
\midrule
Teacher(GPT)  & 71.2 & 68.7 & 78.5 & 8.2 & 11.5 & 10.8 & 100.0 & 100.0 & 100.0 & 5.8 & 7.4 & 6.2 \\
\midrule
KD-120M & 37.3 & 34.9 & 46.1 & 13.2 & 16.4 & 15.3 & 57.1 & 52.7 & 63.4 & 9.3 & 10.9 & 9.2 \\
SeqKD-120M & 38.5 & 36.0 & 47.2 & 12.9 & 16.0 & 15.0 & 58.3 & 54.0 & 64.2 & 9.2 & 10.8 & 9.0 \\
Token-120M & 39.4 & 36.7 & 48.3 & 12.4 & 15.7 & 14.8 & 59.3 & 55.1 & 65.7 & 9.1 & 10.7 & 8.9 \\
Ours (120M) & \textbf{43.7} & \textbf{41.2} & \textbf{52.8} & \textbf{11.2} & \textbf{14.6} & \textbf{13.8} & \textbf{62.3} & \textbf{58.7} & \textbf{66.2} & \textbf{8.2} & \textbf{9.5} & \textbf{7.8} \\
\midrule
KD-340M & 49.3 & 47.1 & 58.8 & 11.1 & 14.9 & 13.8 & 66.0 & 60.5 & 68.3 & 8.0 & 9.6 & 8.2 \\
SeqKD-340M & 50.7 & 48.6 & 60.2 & 10.9 & 14.5 & 13.4 & 67.2 & 61.4 & 69.3 & 7.9 & 9.5 & 8.1 \\
Token-340M & 52.1 & 49.8 & 61.5 & 10.6 & 14.1 & 13.0 & 68.1 & 62.4 & 70.5 & 7.8 & 9.4 & 7.9 \\
Ours (340M) & \textbf{56.3} & \textbf{54.7} & \textbf{65.5} & \textbf{9.8} & \textbf{13.1} & \textbf{12.2} & \textbf{71.5} & \textbf{66.9} & \textbf{74.0} & \textbf{7.1} & \textbf{8.7} & \textbf{7.0} \\
\midrule
KD-760M & 57.5 & 54.3 & 66.2 & 10.1 & 13.7 & 12.6 & 72.0 & 67.2 & 73.1 & 7.2 & 8.9 & 7.4 \\
SeqKD-760M & 58.7 & 55.5 & 67.4 & 9.8 & 13.4 & 12.3 & 73.1 & 68.3 & 74.3 & 7.1 & 8.8 & 7.3 \\
Token-760M & 60.2 & 57.0 & 69.1 & 9.5 & 13.2 & 12.0 & 74.0 & 69.3 & 76.4 & 7.0 & 8.6 & 7.2 \\
Ours (760M) & \textbf{64.8} & \textbf{61.5} & \textbf{73.1} & \textbf{8.9} & \textbf{12.4} & \textbf{11.7} & \textbf{77.9} & \textbf{73.1} & \textbf{80.4} & \textbf{6.4} & \textbf{8.1} & \textbf{6.6} \\
\bottomrule
\end{tabular}
}
\vspace{-2pt}
\end{table*}

\begin{table*}[t]  \small
\vspace{-3pt}
\centering
\caption{
Unified comparison of task success (↑), reasoning length (↓), CoT match (↑), and episode latency (↓) across ALFWorld, WebShop, and HotPotQA. The teachers are OPT-13B, LLaMA-13B, and Orca2-13B ReAct-style agents. Students trained via \textbf{Structured Agent Distillation} consistently outperform token-level~\cite{gu2024minillm}, KD~\cite{distilbert,lightpaff}, and SeqKD~\cite{skd,vicuna,alpaca,ITGPT4,lima} baselines.
}
\vspace{-6pt}
\label{tab:opt_llama_scaling_all}
\resizebox{0.8\textwidth}{!}{
\begin{tabular}{l|ccc|ccc|ccc|ccc}
\toprule
\textbf{Model} & \multicolumn{3}{c|}{\textbf{Task Success ↑}} & \multicolumn{3}{c|}{\textbf{Reasoning Length ↓}} & \multicolumn{3}{c|}{\textbf{CoT Match Rate↑}} & \multicolumn{3}{c}{\textbf{Episode Latency ↓}} \\
& ALF & Web & Hot & ALF & Web & Hot & ALF & Web & Hot & ALF & Web & Hot \\
\midrule
Teacher (OPT13B) & 76.5 & 73.2 & 82.7 & 38.2 & 35.9 & 40.7 & 100.0 & 100.0 & 100.0 & 6.5 & 5.9 & 4.8 \\
\midrule
KD (OPT-1.3B) & 45.3 & 40.7 & 51.0 & 47.6 & 43.9 & 50.1 & 58.9 & 55.0 & 67.3 & 8.1 & 7.3 & 6.3 \\
SeqKD (OPT-1.3B) & 46.2 & 41.8 & 52.3 & 46.3 & 42.5 & 49.4 & 60.2 & 56.1 & 68.4 & 7.9 & 7.2 & 6.2 \\
Token-OPT-1.3B & 47.8 & 43.2 & 54.1 & 45.7 & 41.8 & 48.5 & 61.5 & 57.9 & 69.8 & 7.8 & 7.1 & 6.0 \\
Ours (OPT-1.3B) & \textbf{52.3} & \textbf{48.7} & \textbf{58.5} & \textbf{41.2} & \textbf{38.0} & \textbf{43.6} & \textbf{67.2} & \textbf{63.8} & \textbf{74.4} & \textbf{7.0} & \textbf{6.4} & \textbf{5.3} \\
\midrule
KD (OPT-2.7B) & 53.1 & 48.3 & 60.7 & 44.3 & 40.1 & 46.5 & 65.1 & 61.0 & 73.4 & 7.5 & 6.9 & 5.7 \\
SeqKD (OPT-2.7B) & 54.4 & 49.7 & 61.3 & 43.2 & 39.5 & 46.0 & 66.2 & 61.9 & 74.1 & 7.3 & 6.7 & 5.6 \\
Token-OPT-2.7B & 55.6 & 51.0 & 62.9 & 42.5 & 39.0 & 45.2 & 67.3 & 62.7 & 75.5 & 7.2 & 6.6 & 5.6 \\
Ours (OPT-2.7B) & \textbf{59.2} & \textbf{56.4} & \textbf{67.0} & \textbf{39.4} & \textbf{36.2} & \textbf{41.7} & \textbf{71.6} & \textbf{67.9} & \textbf{79.8} & \textbf{6.7} & \textbf{6.1} & \textbf{5.0} \\
\midrule
KD (OPT-6.7B) & 60.1 & 55.8 & 67.2 & 42.1 & 38.4 & 43.9 & 70.1 & 66.5 & 78.1 & 7.0 & 6.4 & 5.4 \\
SeqKD (OPT-6.7B) & 61.3 & 56.9 & 68.1 & 41.4 & 37.8 & 43.3 & 71.4 & 67.0 & 79.0 & 6.9 & 6.3 & 5.4 \\
Token-OPT-6.7B & 62.8 & 58.6 & 69.7 & 40.8 & 37.2 & 42.9 & 72.2 & 67.9 & 80.2 & 6.8 & 6.2 & 5.3 \\
Ours (OPT-6.7B) & \textbf{67.1} & \textbf{63.8} & \textbf{73.9} & \textbf{38.0} & \textbf{35.1} & \textbf{40.2} & \textbf{76.4} & \textbf{72.5} & \textbf{84.0} & \textbf{6.5} & \textbf{5.9} & \textbf{4.9} \\
\midrule
\midrule
Teacher (LLaMA13B) & 75.3 & 71.8 & 81.0 & 37.5 & 34.8 & 39.9 & 100.0 & 100.0 & 100.0 & 6.4 & 5.8 & 4.7 \\
\midrule
KD (LLaMA-7B) & 62.1 & 56.9 & 70.1 & 42.5 & 38.3 & 44.0 & 71.2 & 66.0 & 79.0 & 6.8 & 6.3 & 5.3 \\
SeqKD (LLaMA-7B) & 63.0 & 58.1 & 70.9 & 41.7 & 37.9 & 43.6 & 72.5 & 67.1 & 80.0 & 6.8 & 6.2 & 5.2 \\
Token-LLaMA-7B & 64.2 & 59.3 & 71.5 & 41.1 & 37.5 & 43.2 & 73.0 & 68.3 & 81.2 & 6.7 & 6.1 & 5.2 \\
Ours (LLaMA-7B) & \textbf{68.0} & \textbf{64.1} & \textbf{75.2} & \textbf{38.2} & \textbf{34.9} & \textbf{39.8} & \textbf{77.2} & \textbf{72.9} & \textbf{84.7} & \textbf{6.4} & \textbf{5.8} & \textbf{4.8} \\
\midrule
Teacher (Orca2-13B)& 78.1 & 75.6 & 84.3 & 37.0 & 34.6 & 39.1 & 100.0 & 100.0 & 100.0 & 6.3 & 5.8 & 4.6 \\
\midrule
KD (Orca2-7B) & 64.0 & 59.2 & 72.4 & 41.6 & 37.8 & 43.1 & 73.1 & 68.7 & 81.5 & 6.7 & 6.2 & 5.2 \\
 SeqKD (Orca2-7B) & 65.2 & 60.4 & 73.5 & 40.9 & 37.2 & 42.5 & 74.3 & 69.4 & 82.6 & 6.6 & 6.1 & 5.1 \\
Token-Orca2-7B & 66.3 & 61.7 & 74.8 & 40.3 & 36.9 & 42.0 & 75.6 & 70.2 & 83.4 & 6.6 & 6.0 & 5.1 \\
Ours (Orca2-7B) & \textbf{70.5} & \textbf{66.2} & \textbf{78.6} & \textbf{37.8} & \textbf{34.2} & \textbf{38.9} & \textbf{79.4} & \textbf{74.6} & \textbf{86.5} & \textbf{6.3} & \textbf{5.8} & \textbf{4.7} \\
\bottomrule
\end{tabular}
}
\vspace{-10pt}
\end{table*}

\noindent\textbf{Task Success Rate.}
As shown in Table~\ref{tab:all_metrics_combined}, \textbf{Structured Agent Distillation} consistently outperforms token-level MiniLLM baselines across all student sizes, with especially notable gains at 120M (+4.3\%), confirming the effectiveness of trajectory-level supervision over token imitation.\\
\noindent\textbf{Reasoning Efficiency.}
Table~\ref{tab:all_metrics_combined} shows that students trained via \textbf{Structured Agent Distillation} generate shorter reasoning spans.\\
\noindent\textbf{CoT Consistency (defined in Appendix~\ref{appendix:metrics}).}
As shown in Table~\ref{tab:all_metrics_combined}, our method achieves higher CoT match rates across all settings, demonstrating stronger structural alignment with teacher reasoning.\\
\noindent\textbf{Latency.}
Following prior work~\cite{yao2023react, shinn2023reflexion}, we measure the average number of reasoning and action steps per episode. As shown in Table~\ref{tab:all_metrics_combined}, \textbf{segment-aware} students consistently exhibit shorter execution traces. Latency is measured in reasoning/action steps rather than wall-clock time, and lower values indicate more concise and efficient decision-making.\\
\noindent\textbf{Segment Mask Validation.}
Figure~\ref{fig:mask_overlay_triad} (Appendix F) confirms that token-level masks align accurately with reasoning/action spans across environments.\\   
\noindent\textbf{Span Statistics.}
As shown in Figure~\ref{fig:span_statistics}, reasoning spans are longer and more variable, while action spans are shorter—justifying span-specific supervision.\\
\noindent\textbf{Summary.}
Our method outperforms token-level distillation across all benchmarks (Table~\ref{tab:opt_llama_scaling_all}), yielding more accurate and faithful student agents.

\section{Scaling Analysis}
\label{sec:scaling}
\begin{figure}[t]
\centering
\includegraphics[width=1.0\linewidth]{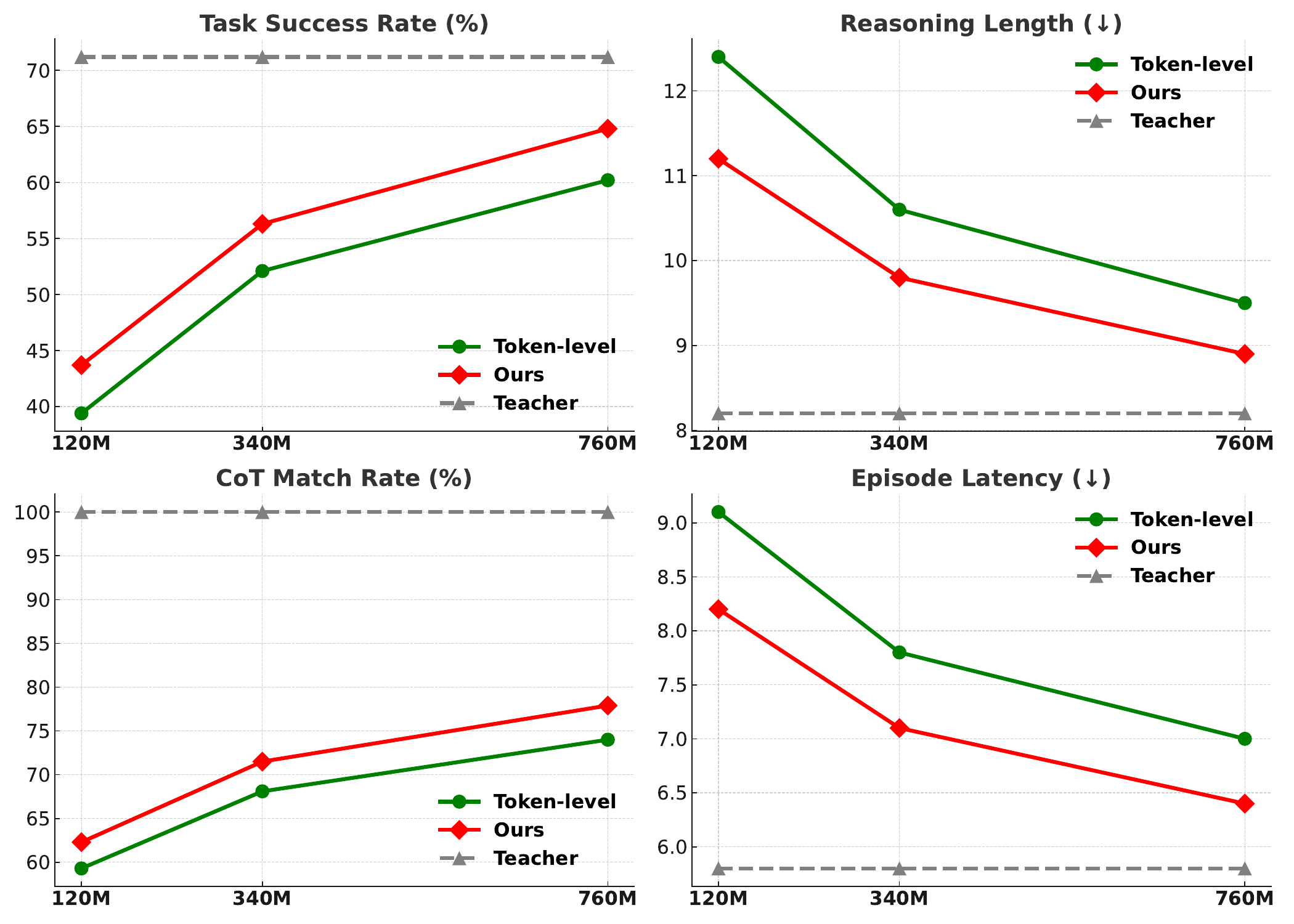}
\vspace{-10pt}
\caption{
Scaling behavior of student agents across model sizes.
\textbf{Top-Left:} Task Success Rate (\%).
\textbf{Top-Right:} Reasoning Length (tokens).
\textbf{Bottom-Left:} Chain-of-Thought (CoT) Match Rate (\%).
\textbf{Bottom-Right:} Episode Latency (steps).
\textbf{Structured Agent Distillation} consistently outperforms the best baseline method (Token-level~\cite{gu2024minillm}) and better approaches teacher performance as model capacity increases.
}
\vspace{-8pt}
\label{fig:scaling_analysis}
\Description{ \textbf{Scaling behavior of student agents across model sizes.}}
\end{figure}

To assess scalability, we transfer trajectories from a GPT-2-1.5B teacher model into student models with 120M, 340M, and 760M parameters using \textbf{Structured Agent Distillation}, and compare the outcomes against the best-performing baseline method (token-level~\cite{gu2024minillm}).

Figure~\ref{fig:scaling_analysis} summarizes four metrics—task success rate, reasoning efficiency (avg. reasoning length), CoT match rate, and episode latency—on ALFWorld, WebShop, and HotPotQA-ReAct.\\
\noindent\textbf{Task Success.}
Success rates improve with model size (top-left), with students trained via \textbf{Structured Agent Distillation} consistently outperforming token-level baselines. At 760M, performance closely approaches the teacher.\\
\noindent\textbf{Reasoning Efficiency.}
Students trained via \textbf{Structured Agent Distillation} produce shorter, more efficient reasoning traces (top-right), especially at larger scales.\\
\noindent\textbf{CoT Match.}
Students distilled with our method better recover the teacher’s reasoning structure (bottom-left), with consistently higher CoT match rates.\\
\noindent\textbf{Latency.}
Structured supervision yields lower episode latency (bottom-right), reducing decision steps and accelerating task completion.\\
\textbf{Structured Agent Distillation} scales effectively with student capacity, enhancing task success, planning efficiency, and structural reasoning alignment.
Improvements are most pronounced at smaller scales (e.g., 120M, 340M), where it mitigates the performance degradation commonly observed with token-level imitation.
Additional scaling results for OPT and LLaMA models are presented in Appendix G. 

\section{Ablation Studies}
\label{sec:ablation}

We conduct ablation studies to understand the contribution of each component in our \textbf{Structured Agent Distillation} framework.
\subsection{Ablation on Reasoning, Action, and Segmentation Components}
Specifically, we analyze the roles of reasoning supervision, action supervision, and span segmentation.

\begin{table*}[ht] 
\centering
\caption{
Ablation study on ALFWorld using a 340M-parameter student model. 
Each variant controls whether reasoning, action supervision, and explicit span segmentation are enabled. 
We report task success rate, CoT match rate, and episode latency(steps).
}
\label{tab:ablation_combined}
\begin{tabular}{lccc|ccc}
\toprule
\textbf{Method} & \textbf{Reason} & \textbf{Action} & \textbf{Segm} & \textbf{Succ (\%)} ↑ & \textbf{CoT (\%)} ↑ & \textbf{Episode Latency} ↓ \\
\midrule
Full Segment-Aware (Ours) & \cmark & \cmark & \cmark & \textbf{56.3} & \textbf{71.5} & \textbf{7.1} \\
Only Reasoning Supervision    & \cmark &  \textcolor{red}{\ding{55}} & \cmark & 52.7 & 69.2 & 7.4 \\
Only Action Supervision       & \textcolor{red}{\ding{55}} & \cmark & \cmark & 49.5 & 61.8 & 7.8 \\
No Span Segmentation          & \cmark & \cmark &  \textcolor{red}{\ding{55}} & 48.2 & 60.4 & 8.1 \\
Random Span Masking           & \cmark & \cmark & $\sim$ (random) & 45.9 & 57.7 & 8.4 \\
\bottomrule
\end{tabular}
\vspace{-10pt}
\end{table*}

Table~\ref{tab:ablation_combined} confirms that each supervision component plays a critical role:\\
\noindent\textbf{Removing reasoning supervision} ($\mathcal{L}_{\text{CoT}}$) significantly degrades CoT matches and increases latency, indicating reduced planning coherence. \\
\noindent\textbf{Removing action supervision} ($\mathcal{L}_{\text{Act}}$) lowers task success and execution fidelity, showing the importance of behavior alignment. \\
\noindent\textbf{Disabling span segmentation} (flat token-level loss) causes uniform degradation across all metrics, suggesting that structural decomposition is essential. \\
\noindent\textbf{Random span masking} further harms performance, highlighting the need for accurate, semantically meaningful segmentation.

These results demonstrate that our method 
captures structurally distinct signals crucial for reasoning-action alignment.

\subsection{Ablation on \texorpdfstring{$\lambda_r$:$\lambda_a$}{lambda	extunderscore r:lambda	extunderscore a} Ratio}
\label{sec:ablation_cot_act_ratio1}
We conduct an ablation study to investigate how the balance between reasoning and action supervision affects student performance.
Specifically, we vary the weighting ratio between the CoT loss ($\lambda_{\text{CoT}}$) and the action loss ($\lambda_{\text{Act}}$), while keeping the total loss weight fixed.
As shown in Table~\ref{tab:ablation_cot_act_ratio1}, using both types of supervision yields the best results, with a balanced 1:1 ratio outperforming all other settings across task success rate, CoT match rate, and episode length.
Interestingly, even when using only one type of supervision (either $\lambda_{\text{CoT}}$ or $\lambda_{\text{Act}}$ set to zero), our structured distillation approach still outperforms token-level baselines.
These results highlight the complementary nature of reasoning and action spans in learning efficient, high-quality student agents.

\begin{table}[h]
\centering
\small 
\setlength{\tabcolsep}{4pt} 
\caption{ 
Ablation study on ALFWorld using a 340M-parameter student model.  
We vary the loss weight ratios between reasoning (CoT) and action supervision 
while keeping their sum fixed.  
Structured Agent Distillation consistently outperforms token-level baselines across all settings.
}
\label{tab:ablation_cot_act_ratio1}
\vspace{-5pt}
\begin{tabular}{@{}c|ccc@{}}
\toprule
\textbf{CoT : Act Ratio} & \textbf{Succ (\%)} ↑ & \textbf{CoT  (\%)} ↑ & \textbf{Episode Latency} ↓ \\
\midrule
1.0 : 0.0 & 53.4 & 69.2 & 8.2 \\
0.0 : 1.0 & 52.7 & 66.1 & 8.3 \\
0.5 : 1.0 & 55.8 & 70.3 & 7.7 \\
1.0 : 1.0 & \textbf{56.3} & \textbf{71.5} & \textbf{7.1} \\
2.0 : 1.0 & 56.1 & 70.8 & 7.3 \\
\midrule
Token-Level Baseline & 52.1 & 68.1 & 8.6 \\
\bottomrule
\end{tabular}
\end{table}

Ablation study on curriculum sampling in Appendix~\ref{appendix:curriculum},
Qualitative Examples and Faithfulness in Appendix~\ref{app:sec:qualitative}.
\section{Discussion}
\label{sec:discussion}
\noindent\textbf{Semantic Decoupling Matters.}
Our results show that simply applying token-level imitation—even with large-scale training—fails to preserve the structure of reasoning-action workflows. By explicitly segmenting trajectories and supervising each span separately, \textbf{Structured Agent Distillation} enables student models to independently learn symbolic reasoning and grounded execution behaviors. This semantic decoupling proves essential, especially under limited capacity.\\
\noindent\textbf{Loss Design and Supervision.}
Although the total loss combines $\mathcal{L}_{\text{CoT}}$ and $\mathcal{L}_{\text{Act}}$, ablation studies show that removing either term causes clear degradation in CoT alignment, task success, and planning efficiency.
This demonstrates that span-level supervision offers complementary signals beyond token-level KL objectives.

\noindent\textbf{Robustness of Rule-Based Segmentation.}
Our segmentation pipeline relies on task-specific patterns to extract \texttt{[REASON]} and \texttt{[ACT]} spans. While simple, this rule-based approach is highly reliable across benchmarks and avoids the cost of manual annotation. Empirically, segmentation accuracy is sufficient to yield clear improvements in student performance.\\

\noindent\textbf{Scalability, Generalization, and Frontier LLMs.}
Due to the resource demands of ultra-large models, we adopt 13B-scale teachers (OPT, LLaMA, Orca2). 
{Structured Agent Distillation (SAD)} is inherently \emph{scaling- and model-agnostic}. 
Its span-level supervision functions independently of the architecture, tokenizer, 
and consistent gains across both decoder-only and instruction-tuned families 
validate its architectural generality. 
Moreover, SAD extends naturally to frontier LLMs such as {Qwen3-235B-A22B} and {DeepSeek-V3.1}, 
where distributed H100 inference or API-based frozen-logit distillation enables 235B$\!\rightarrow\!$13B/7B compression. 
This extension preserves the same structured objective, confirming that SAD generalizes to ultra-large LLMs without modification.

\section{Related Work}

\noindent\textbf{LLM Agents.}
LLM agents unify reasoning and action. ReAct~\cite{yao2023react} interleaves language and actions. Toolformer~\cite{schick2023toolformer} self-supervises API calls, while WebGPT~\cite{nakano2021webgpt} reasons via web browsing. AutoGPT~\cite{autogpt2023} chains subtasks autonomously. AgentBench~\cite{liu2023agentbench}, ReWoo~\cite{xu2023rewoo}, and HuggingGPT~\cite{shen2023hugginggpt} explore modularity and tool use. CAMEL~\cite{li2023camel}, ChatDev~\cite{chatdev}, AutoGen~\cite{wu2023autogen}, CrewAI~\cite{crewai_github}, ToolBench~\cite{guo2024stabletoolbench}, and LangChain~\cite{langchain_docs} focus on multi-agent collaboration and LLM orchestration. GITM~\cite{zhu2023ghost} integrates LLMs with memory and knowledge to build generally capable agents.
Although these agents show strong reasoning-action abilities, their trajectories remain challenging to compress and generalize.

\noindent\textbf{Distillation and Fine-Tuning.}
Token-level distillation~\cite{hinton2015distilling, eldan2023tinystories, gu2024minillm,minixhofer2025cross,cui2025multi,zhang2024knowledge} compresses LLMs via soft target alignment and teacher-guided training. Recent work improves multi-granular supervision~\cite{gu2024minillm}, cross-modal transfer~\cite{zhou2023unidistill}, and compact pretraining~\cite{zhang2024tinyllama}. However, these methods target static text generation, not structured reasoning-action trajectories.

\noindent\textbf{Trajectory Modeling and Behavioral Cloning.}
Recent work in sequence-level imitation~\cite{wang2022self,ke2023ccil,swamy2022sequence,cundy2023sequencematch,le2016smooth,yang2023seq2seq,zhang2020nested}, behavior cloning~\cite{codevilla2018end,hepburn2024model,zhou2024developing,wen2020fighting,torabi2018behavioral,robertson2020concurrent} emphasizes temporal coherence in agent behavior. SpanBERT~\cite{joshi2020spanbert} masks contiguous spans instead of individual tokens. 
ConvBERT~\cite{jiang2020convbert} uses span-based convolutions instead of attention heads. 
LLM agents require structured objectives to model reasoning and action jointly—captured by our \textbf{Structured Agent Distillation} framework.

\FloatBarrier  
\section{Conclusion}
\label{sec:conclusion}

We present \textbf{Structured Agent Distillation}, a compression framework that segments teacher trajectories into reasoning and action spans, allowing students to better replicate high-level reasoning and low-level execution beyond token-level imitation.
Our method achieves consistent gains in task success, reasoning efficiency, and CoT alignment over token-level baselines. Scaling and ablation results confirm the benefits of structured supervision under limited capacity.
These findings highlight the importance of preserving trajectory structure for training lightweight agents and open avenues for structured knowledge transfer in real-world decision making.

\clearpage  



\bibliographystyle{ACM-Reference-Format} 
\bibliography{sample}



\newpage
\clearpage
\onecolumn
\appendix

\section*{Appendix}
\section{ Structured Agent Distillation framework}
\label{appendix:framework}
\subsection{Teacher-Student Interaction Illustration}

\begin{figure}[h]
    \centering
    \includegraphics[width=0.6\textwidth]{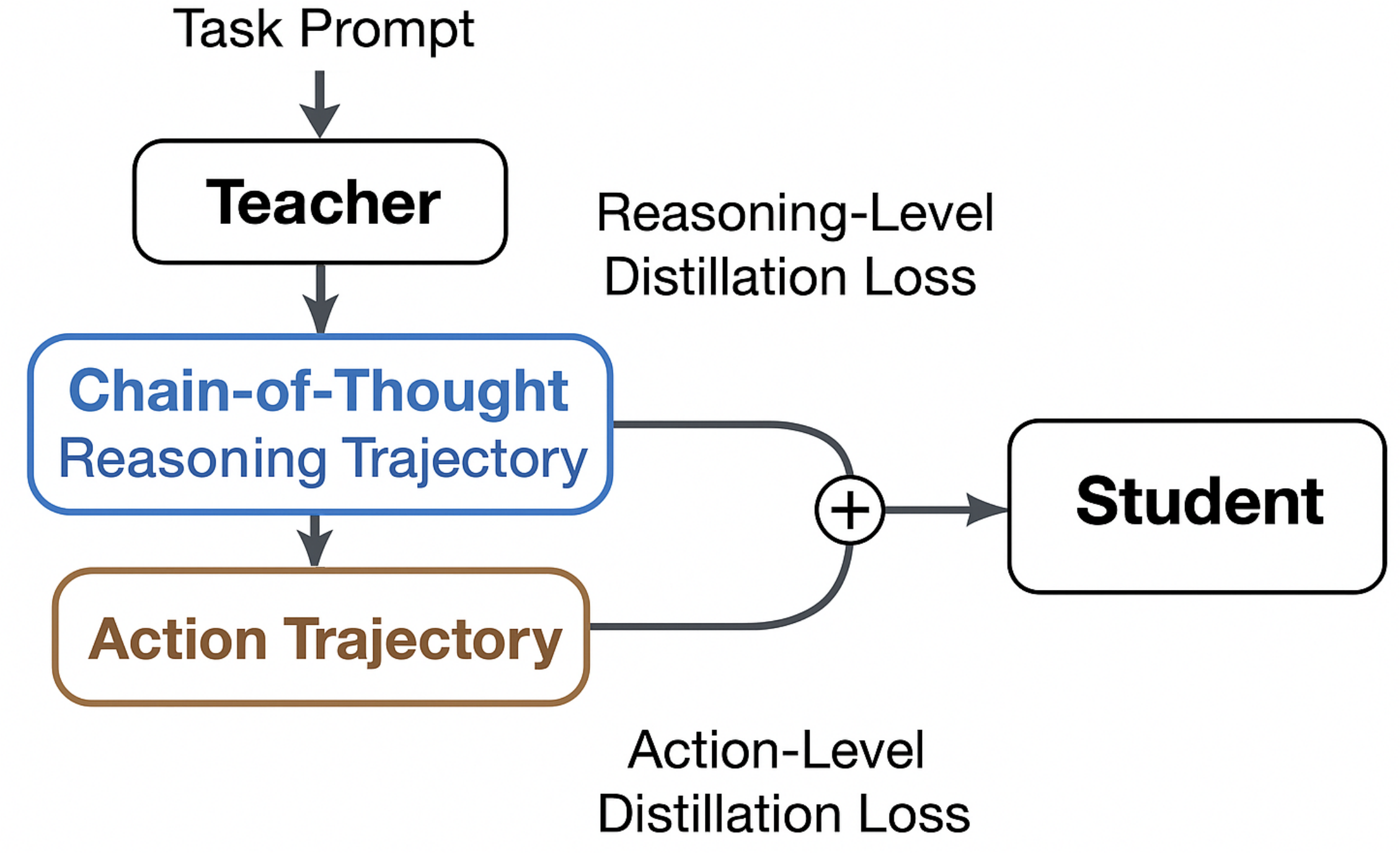}
    \caption{
        Illustration of the teacher-student interaction in our \textbf{Structured Agent Distillation} framework.
        The teacher model generates both the chain-of-thought (CoT)~\cite{ott2023thoughtsource} reasoning steps and the corresponding action trajectory based on the task input.
        The student model learns to imitate these trajectories by minimizing both reasoning-level and action-level distillation losses.
        This joint supervision allows the student to capture both high-level problem-solving strategies and low-level task executions.
    }
    \label{fig:teacher_student_interaction}
    \Description{An overview of the loss flow in our \textbf{Structured Agent Distillation} framework. The student model learns from the teacher's trajectory by decomposing the learning signals into reasoning and action objectives. These loss terms are aggregated to guide effective student optimization.}
\end{figure}
Figure~\ref{fig:teacher_student_interaction} illustrates the core mechanism of our \textit{Structured Agent Distillation} framework. Given a task prompt, the teacher model first produces a \textbf{chain-of-thought reasoning trajectory}, which captures its intermediate thinking steps and decision rationale. It then generates an \textbf{action trajectory}, consisting of task-specific outputs (e.g., tool invocations, API calls, environment actions).

The student model receives both trajectories as supervision signals. By minimizing a combination of reasoning-level and action-level distillation losses, it learns to emulate not only the output behavior but also the underlying reasoning patterns of the teacher. This two-level guidance enables the student to generalize better across reasoning-intensive agent tasks, even with fewer parameters.

\subsection{Reasoning–Action Segmentation}
\begin{figure}[t]
  \centering
  \includegraphics[width=0.4\linewidth]{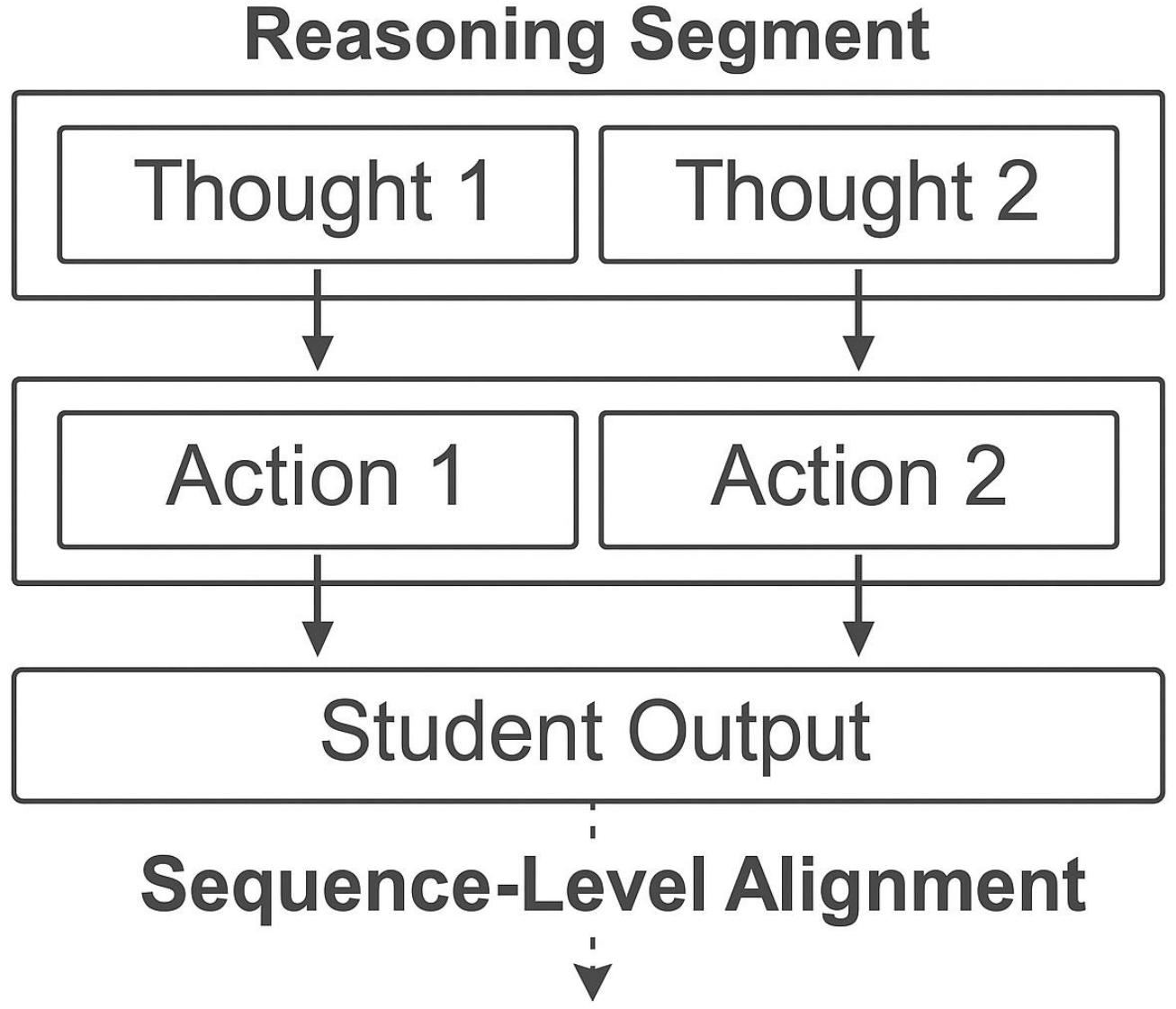}
 \caption{Illustration of the segmented trajectory structure used in our \textbf{Structured Agent Distillation} framework. The teacher’s trajectory is divided into a reasoning segment (language-based, multi-step inference) and an action segment (tool invocation or final structured decision). Each segment is used to compute a dedicated imitation loss for the student model.}

  \label{fig:segment}
  \Description{Illustration of the segmented trajectory structure used in our \textbf{Structured Agent Distillation} framework. The teacher’s trajectory is divided into a reasoning segment (language-based, multi-step inference) and an action segment (tool invocation or final structured decision). Each segment is used to compute a dedicated imitation loss for the student model.}
\end{figure}
To better supervise the distillation process, we segment each demonstration trajectory into two interpretable components: the \textbf{Reasoning Segment} and the \textbf{Action Segment}, as shown in Figure~\ref{fig:segment}.

The \textbf{Reasoning Segment} contains the teacher's natural language chain-of-thoughts that progressively unfold the problem-solving logic, while the \textbf{Action Segment} consists of final decisions, structured commands, or tool invocations derived from the reasoning.

During training, we apply separate segment masks \( m_r \) and \( m_a \) to isolate these parts. This enables the student model to be supervised with segment-specific objectives: a chain-of-thought imitation loss \( \mathcal{L}_{\text{cot}} \), an action prediction loss \( \mathcal{L}_{\text{act}} \), and a trajectory-level consistency loss \( \mathcal{L}_{\text{traj}} \).

This design allows for fine-grained control over the learning process, encouraging the student to mimic both the rationale and the resulting behavior.

\subsection{Loss Flow in Structured Agent Distillation}

\begin{figure}[h]
\centering
\includegraphics[width=0.4\linewidth]{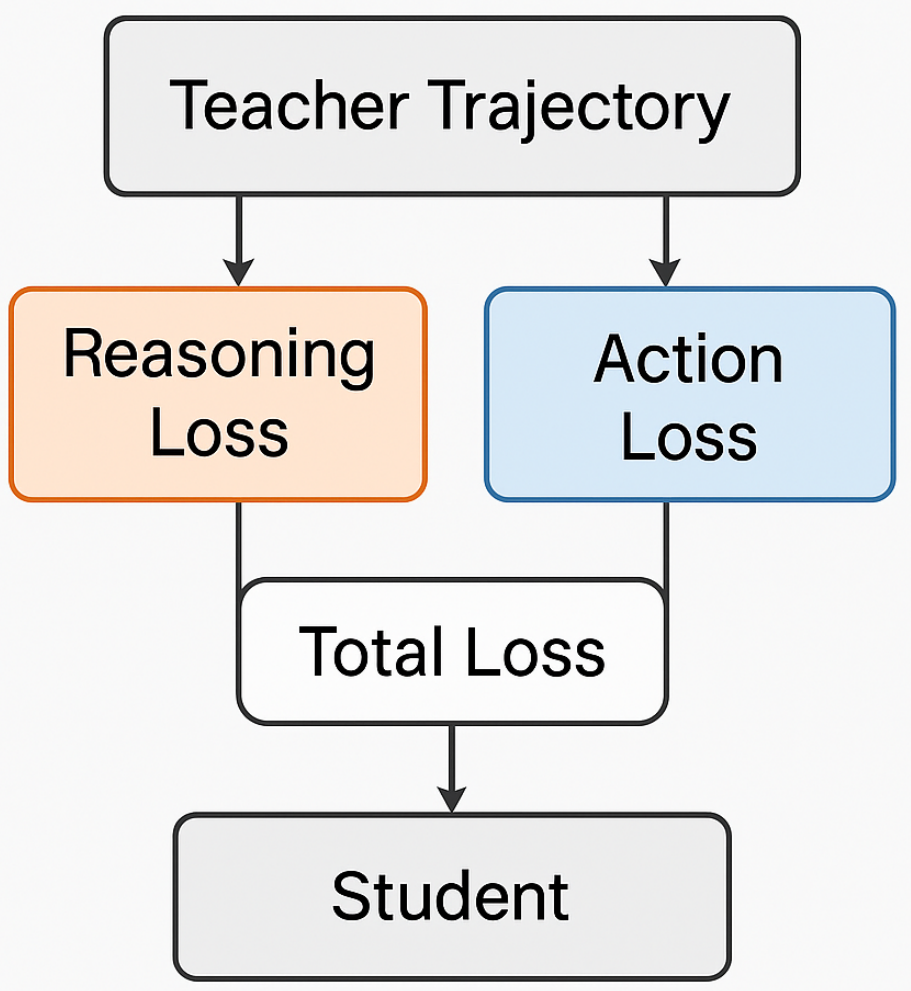}
\caption{An overview of the loss flow in our \textbf{Structured Agent Distillation} framework. The student model learns from the teacher's trajectory by decomposing the learning signals into reasoning and action objectives. These loss terms are aggregated to guide effective student optimization.}

\label{fig:loss_flow}
\Description{An overview of the loss flow in our \textbf{Structured Agent Distillation} framework. The student model learns from the teacher's trajectory by decomposing the learning signals into reasoning and action objectives. These loss terms are aggregated to guide effective student optimization.}
\end{figure}
Figure~\ref{fig:loss_flow} illustrates how different loss components are computed and used to optimize the student model. The input to the student model is a trajectory consisting of both reasoning and action segments generated by the teacher model. Segment masks are applied to distinguish the two parts.

Two types of loss are computed:
\begin{itemize}
    \item \textbf{Reasoning Loss} \( \mathcal{L}_{\text{cot}} \): aligns the student's chain-of-thought outputs with the teacher's intermediate reasoning steps.
    \item \textbf{Action Loss} \( \mathcal{L}_{\text{act}} \): ensures that the student reproduces the teacher's final decision or action.
\end{itemize}

These loss terms are weighted and aggregated as:
\[
\mathcal{L}_{\text{total}} = \lambda_1 \mathcal{L}_{\text{cot}} + \lambda_2 \mathcal{L}_{\text{act}}   
\]
which is then used for backpropagation to update the student model parameters.

\subsection{Curriculum Sampling Process}

\begin{figure}[!t]
\centering
\includegraphics[height=0.4\textheight, keepaspectratio]{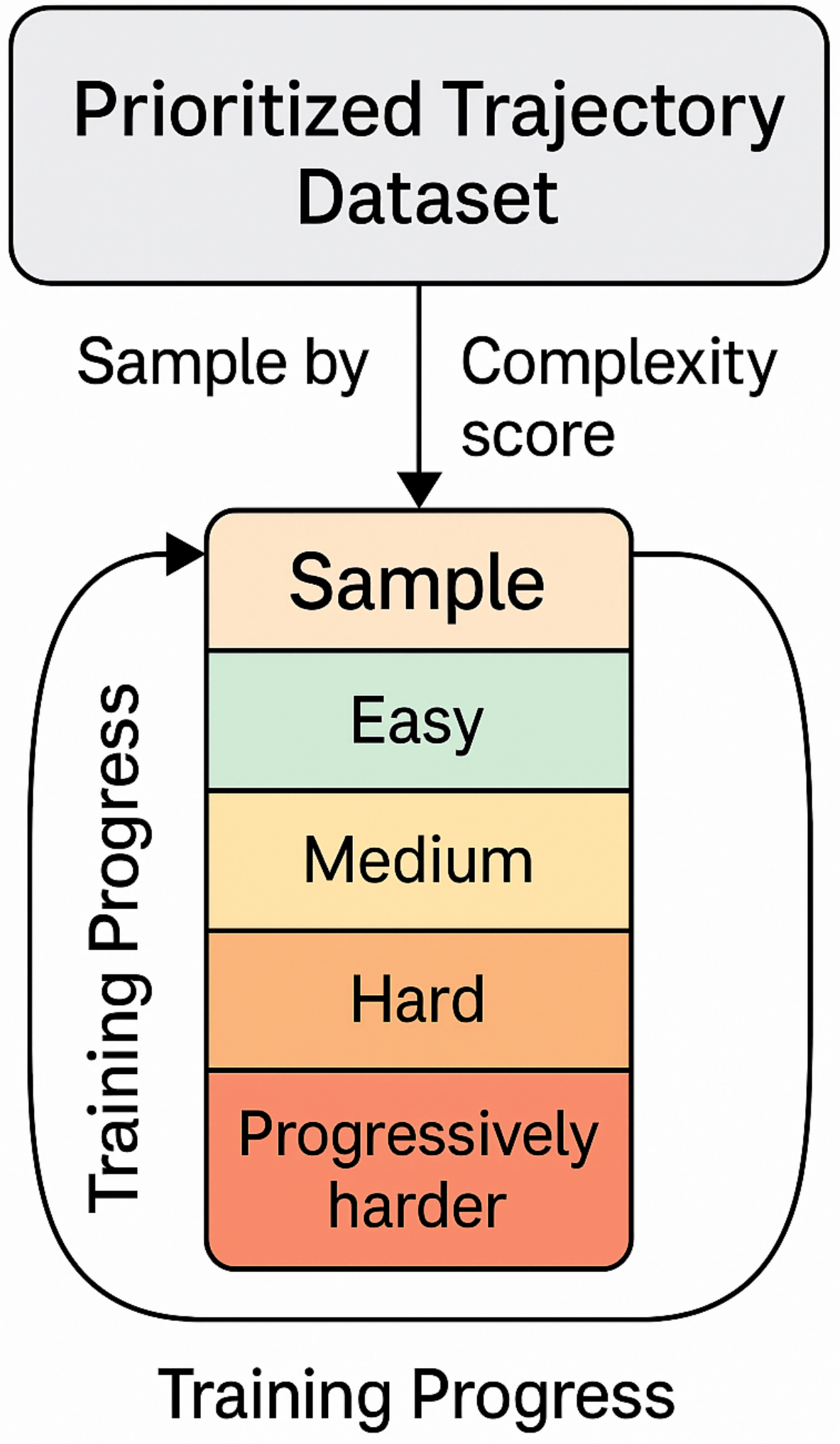}
\caption{Curriculum Sampling Process.  
The scheduler gradually increases task difficulty during training,
guiding the student model from simple to complex structured agent trajectories
under Structured Agent Distillation.}

\label{fig:curriculum-sampling}
 \Description{Illustration of the curriculum sampling process. The scheduler gradually increases task difficulty during training, guiding the student model from simple to complex \textbf{structured agent trajectories} under \textbf{Structured Agent Distillation}.}
\end{figure}

In Figure~\ref{fig:curriculum-sampling}, to guide the student model's learning in a progressive manner, we apply a curriculum learning strategy. The curriculum scheduler \( \mathcal{C} \) starts with easier reasoning-action trajectories and gradually introduces harder samples as training progresses. This improves training stability and enhances generalization, especially in long-horizon tasks.

\section{Task and Environment Setup}
\label{appendix:task-setup}

\noindent\textbf{ALFWorld.}
ALFWorld~\cite{ALFWorld20} is an embodied instruction-following environment where agents must complete household tasks through navigation and object manipulation.
Each episode provides a natural language goal (e.g., "Put the soap on the sink"), and the agent interacts with a simulated environment to achieve it.
We collect teacher trajectories from a ReAct-style agent that interleaves reasoning and actions, and segment them into reasoning spans and action spans.

\noindent\textbf{WebShop.}
WebShop~\cite{yao2022webshop} is a text-based e-commerce environment where agents must fulfill shopping goals through search, click, and buy tool invocations.
Teacher agents generate structured trajectories consisting of reasoning steps (e.g., product comparisons) and concrete tool actions.
Students are trained on these structured trajectories under curriculum sampling.

\noindent\textbf{HotPotQA-ReAct.}
We adapt HotPotQA~\cite{yao2023react} into a ReAct-style question answering setup, where teacher agents interleave multi-hop reasoning steps before final answers.
Trajectories are segmented into reasoning and action spans, ensuring a clear separation between intermediate thoughts and conclusive outputs.
Students are trained to imitate these structured trajectories under segment-aware supervision.

\noindent\textbf{Models.}
We apply \textbf{Structured Agent Distillation} to compress ReAct-style GPT-2-1.5B teacher trajectories into student models of 120M, 340M, and 760M (GPT-2); OPT-13B into OPT students (1.3B, 2.7B, 6.7B); and LLaMA-13B into LLaMA-7B, Orca2-13B into Orca2-7B. \\
Trajectories are generated offline via low-temperature ReAct prompts, segmented into reasoning and action spans (Appendix~\ref{appendix:data}), and used to train students with segment-aware losses.

\noindent\textbf{Metrics.}
We evaluate student agents on: 
\textbf{(i)} task success rate~\cite{liu2023agentbench},
\textbf{(ii)} average reasoning length (efficiency),
\textbf{(iii)} CoT match rate (consistency), and 
\textbf{(iv)} latency (reasoning + action steps).
Metric definitions are in Appendix~\ref{appendix:metrics}.

\noindent\textbf{Baselines.}
We compare against token-level baselines, primarily using MiniLLM~\cite{gu2024minillm}—a state-of-the-art student model distilled via token-level imitation, which minimizes the KL divergence between student and teacher outputs at every decoding step.\textbf{KD}~\cite{distilbert,lightpaff} fine-tunes the student model on $\mathcal{D}$ using the teacher distribution as the supervision at each token step, also known as word-level KD. \textbf{SeqKD}~\citep{skd,vicuna,alpaca,ITGPT4,lima} fine-tunes the student model on the data generated by the teacher model.

\noindent\textbf{Training Details}
All student agents are initialized 
for ALFWorld and  WebShop. The models are fine-tuned on the collected teacher trajectories using the AdamW optimizer with the following settings:

\begin{itemize}
    \item \textbf{Batch size}: 64 
    \item \textbf{Learning rate}: $2 \times 10^{-5}$ with linear warmup and cosine decay
    \item \textbf{Max steps}: 20k steps for ALFWorld, 10k for WebShop
    \item \textbf{Gradient clipping}: 1.0
    \item \textbf{Sequence length}: 512 tokens (truncated if longer)
    \item \textbf{Distillation loss}: Cross-entropy for token-level, segment-aware alignment loss for our method, with optional trajectory-level contrastive term (weight $\lambda = 0.3$)
\end{itemize}
To support generalization, we used adapter-based tuning, optional KL regularization, and span-level supervision, which helps reduce overfitting. The details are in Appendix J.

The teacher is queried online to avoid distribution drift: offline trajectories become stale as the student deviates from the teacher’s rollout policy, reducing success rates by 2--3\%. In practice, cached teacher logits make online distillation efficient with negligible overhead.

\noindent\textbf{Inference Details}
We followed standard evaluation practice by running each method on 3 independent random seeds per task, using fixed train/test splits. Results are reported as the mean across runs; standard deviation was negligible and omitted for clarity.
All models are trained on A6000 GPU, and evaluation is conducted using deterministic greedy decoding unless otherwise noted.

\section{Evaluation Metrics}
\label{appendix:metrics}

We evaluate models along three axes:
\begin{itemize}
    \item \textbf{Task Success Rate (TSR)~\cite{liu2023agentbench}}:  
    Measures the success ratio of tasks completed by the agent:
    \begin{equation}
        \text{TSR} = \frac{N_{\text{successful tasks}}}{N_{\text{total tasks}}} \times 100\%
    \end{equation}
    where \(N_{\text{successful tasks}}\) is the number of tasks where the final environment state satisfies the success conditions.

    \item \textbf{Reasoning Efficiency (Average Reasoning Length).}
We compute the average number of tokens generated in the reasoning span of each agent trajectory.
Formally, for a set of $N$ trajectories $\{ \tau^{(i)} \}_{i=1}^N$, the average reasoning length (ARL) is defined as:

     \begin{equation}
       \text{ARL} = \frac{1}{N} \sum_{i=1}^{N} \text{Length}(\tau^{(r,i)}),
    \end{equation}

where $\tau^{(r,i)}$ denotes the reasoning span of trajectory $\tau^{(i)}$.
This metric captures the planning efficiency of student agents, with shorter reasoning spans indicating more concise decision-making.

For instance, given three trajectories with reasoning spans of 12, 15, and 14 tokens respectively, the average reasoning length is:

\begin{equation}
\text{ARL} = \frac{12 + 15 + 14}{3} = 13.67.
\end{equation}

   \item \textbf{Chain-of-Thought (CoT) Consistency(CoT Match Rate)}
We evaluate how faithfully the student reproduces the teacher’s reasoning trace.
Specifically, we compute the token-level overlap between the student’s generated reasoning span $\tau_S^{(r,i)}$ and the teacher’s reasoning span $\tau_T^{(r,i)}$, and define CoT match rate as:

\[
\text{CoT Match Rate} = \frac{1}{N} \sum_{i=1}^{N} \frac{|\tau_S^{(r,i)} \cap \tau_T^{(r,i)}|}{|\tau_T^{(r,i)}|},
\]
where  $\cap$ denotes token-level intersection.
Higher CoT match rates indicate better structural fidelity in reasoning reconstruction.\\
This measures the proportion of teacher reasoning tokens reproduced by the student, averaged over $N$ samples.  
Alignment is performed at the token level, ensuring robustness to minor length variation but not to paraphrasing.  

\textbf{Limitations.}  
CoT Match captures lexical overlap rather than semantic equivalence; hence, logically consistent but rephrased reasoning may be undervalued.  
We discuss this limitation in the revision and note that future extensions (e.g., BERTScore-based semantic matching) can complement the metric.

For instance, if a teacher reasoning span contains 9 tokens and a student's reasoning span overlaps with 6 of them, the CoT match rate for that trajectory is:

\[
\text{CoT Match Rate} = \frac{6}{9} = 66.67\%.
\]

\item \textbf{Latency Results (Average Reasoning Steps)~\cite{jin2024impact}.}
As a proxy for agent latency, we measure the average number of generation steps required to complete the reasoning and action phases.
Given $N$ trajectories:

\begin{equation}
\text{Average Steps} = \frac{1}{N} \sum_{i=1}^{N} \text{Steps}(\tau^{(i)}),
\end{equation}
where $\text{Steps}(\tau^{(i)})$ counts the total number of generation steps (tokens or API calls) in trajectory $\tau^{(i)}$.
Lower average steps correspond to faster decision-making and reduced i

For instance, given three sampled trajectories with total generation steps of 16, 13, and 20 respectively, the average number of steps is:

\[
\text{Average Steps} = \frac{16 + 13 + 20}{3} = 16.33.
\]
\end{itemize}

\section{Data Construction for Structured Agent Distillation}
\label{appendix:data}
\subsection{Trajectory Segmentation}
\label{appendix:trajectory_segmentation}
To support \textbf{Structured Agent Distillation}, we decompose each teacher trajectory $\tau$ into two disjoint spans: a reasoning segment $\tau^{(r)}$ and an action segment $\tau^{(a)}$.
This decomposition enables span-specific supervision over the student model’s reasoning and decision behaviors.
Segmentation is performed using lightweight rule-based parsing tailored to each environment.
Specifically:
\begin{itemize}
    \item For \textbf{ALFWorld} and \textbf{WebShop}, each trajectory alternates between lines prefixed with \texttt{Reasoning:} and \texttt{Action:}.
    We extract all lines beginning with \texttt{Reasoning:} as the reasoning span $\tau^{(r)}$, and lines beginning with \texttt{Action:} as the action span $\tau^{(a)}$.
    
    \item For \textbf{HotPotQA-ReAct}, we adopt a two-phase segmentation: text preceding the final \texttt{Answer:} line is assigned to the reasoning span $\tau^{(r)}$, and the \texttt{Answer:} line itself is treated as the action span $\tau^{(a)}$.
\end{itemize}

The segmentation rules across all environments are summarized in Table~\ref{tab:segmentation_rules}.

\begin{table}[h]
\centering
\caption{Segmentation rules for splitting teacher trajectories into reasoning and action spans.}
\label{tab:segmentation_rules}
\begin{tabular}{lcc}
\toprule
\textbf{Environment} & \textbf{Reasoning Span} & \textbf{Action Span} \\
\midrule
ALFWorld & Lines starting with \texttt{Reasoning:} & Lines starting with \texttt{Action:} \\
WebShop & Lines starting with \texttt{Reasoning:} & Lines starting with \texttt{Action:} \\
HotPotQA-ReAct & Lines starting with \texttt{Reasoning:} & Final \texttt{Answer:} line \\
\bottomrule
\end{tabular}
\end{table}

We implement this segmentation using simple regular-expression-based pattern matching.
For instance, the following patterns are used:

\begin{verbatim}
REASONING_REGEX = r"^Reasoning:\s*(.+)"
ACTION_REGEX    = r"^Action:\s*(.+)"
ANSWER_REGEX    = r"^Answer:\s*(.+)"  % for HotPotQA only
\end{verbatim}

We concatenate all lines matching \texttt{Reasoning:} as the reasoning span $\tau^{(r)}$, and all \texttt{Action:}/\texttt{Answer:} lines as the action span $\tau^{(a)}$.
This rule-based segmentation is high-precision and requires no manual annotation.

\subsection{Segment-Aware Token Mask Construction}
\label{appendix:mask_assignment}

After segmenting each trajectory into reasoning and action spans, we construct binary token-level masks to enable selective supervision during \textbf{Structured Agent Distillation}.

Let $x_{1:T}$ denote the tokenized trajectory.

We generate two binary masks:
\begin{itemize}
    \item Reasoning mask $m_r(t) \in \{0,1\}$, where
    \[
    m_r(t) = 
    \begin{cases}
        1 & \text{if token } x_t \text{ belongs to a reasoning span,} \\
        0 & \text{otherwise.}
    \end{cases}
    \]
    \item Action mask $m_a(t) \in \{0,1\}$, where
    \[
    m_a(t) = 
    \begin{cases}
        1 & \text{if token } x_t \text{ belongs to an action span,} \\
        0 & \text{otherwise.}
    \end{cases}
    \]
\end{itemize}

Each token belongs to at most one span, ensuring:
\[
m_r(t) + m_a(t) \leq 1, \quad \forall t \in [1, T].
\]

Tokens outside any supervised span (e.g., prompt headers, formatting artifacts) are excluded from loss computation by masking out their contributions.

\noindent\textbf{Implementation Note.}
We first align the raw text spans to character offsets, then map these to token indices after tokenization.
During training, the distillation losses are selectively applied over tokens using $m_r(t)$ and $m_a(t)$ as weights.

\subsection{Example and Loss Application}
\label{appendix:example_loss}

We illustrate the full segmentation and mask construction process with a concrete example.

\noindent\textbf{Input Example.}
Consider a teacher-generated trajectory excerpt:
\begin{verbatim}
Reasoning: I should find the fridge. It's likely in the kitchen.
Action: goto kitchen
\end{verbatim}

\noindent\textbf{Tokenization and Span Mapping.}
After tokenizing the full sequence (e.g., using the student tokenizer), the token sequence may resemble:
\[
x = [\texttt{Reasoning}, :, \texttt{I}, \texttt{should}, \texttt{find}, \texttt{the}, \texttt{fridge}, ., \texttt{It}, \texttt{'s}, \texttt{likely}, \texttt{in}, \texttt{the}, \texttt{kitchen}, ., \texttt{Action}, :, \texttt{goto}, \texttt{kitchen}]
\]

Based on the span segmentation, we assign masks as follows:
\[
m_r = [1,1,1,1,1,1,1,1,1,1,1,1,1,1,0,0,0,0]
\]
\[
m_a = [0,0,0,0,0,0,0,0,0,0,0,0,0,0,1,1,1,1]
\]

\noindent\textbf{Loss Computation.}
During distillation, we apply span-specific losses weighted by these masks.

When using token-level KL divergence loss:
\begin{equation}
\mathcal{L}_{\text{total}} = 
\lambda_{\text{cot}} \underbrace{\sum_{t=1}^{T} m_r(t) \cdot \text{KL}\left(p_T(x_t) \| p_S(x_t)\right)}_{\mathcal{L}_{\text{cot}}} +
\lambda_{\text{act}} \underbrace{\sum_{t=1}^{T} m_a(t) \cdot \text{KL}\left(p_T(x_t) \| p_S(x_t)\right)}_{\mathcal{L}_{\text{act}}}.
\end{equation}

When using cross-entropy (CE) loss over hard targets:
\begin{equation}
\mathcal{L}_{\text{total}} = 
\lambda_{\text{cot}} \underbrace{\sum_{t=1}^{T} m_r(t) \cdot \text{CE}\left(x_t, \hat{x}_t\right)}_{\mathcal{L}_{\text{cot}}} +
\lambda_{\text{act}} \underbrace{\sum_{t=1}^{T} m_a(t) \cdot \text{CE}\left(x_t, \hat{x}_t\right)}_{\mathcal{L}_{\text{act}}}.
\end{equation}
where $\hat{x}_t$ denotes the student model’s prediction at position $t$.

This segment-aware supervision enables the student agent to imitate both the reasoning trace and the final action decision of the teacher, improving planning fidelity and execution robustness.

\noindent\textbf{Summary.}
This appendix details the full data construction pipeline for \textbf{Structured Agent Distillation}.
We describe how teacher trajectories are segmented into reasoning and action spans, how segment-aware token masks are assigned, and how these masks guide span-specific loss computation.
This \textbf{structured agent supervision} enables fine-grained alignment between student and teacher agents, covering both intermediate reasoning and final action decisions.

\section{Prompt Generation for Structured Agent Distillation}
\label{appendix:prompt}
\subsection{Prompt Templates}
\label{appendix:prompt_tp}
We present the exact prompt formats used in each task during inference and teacher generation.

\noindent\textbf{ALFWorld.}
\texttt{
You are in the [room]. You see: [objects]. What should you do next?}

\noindent\textbf{WebShop.}
\texttt{
User intent: Find a [product type] under [budget]. Available options: [list]. Which item should the user select?
}

\noindent\textbf{HotPotQA-ReAct.}
\texttt{
Q: [question] \\
A: Let's think step by step.
}
We show one representative prompt template per environment used for teacher trajectory generation.

\noindent\textbf{ALFWorld Prompt Example.}
\texttt{
You are an intelligent agent operating in a simulated household environment.  
Goal: Put the soap on the sink.
\\
Reasoning: First, I should find where the soap is located.  
Action: navigate to the bathroom.  
Reasoning: Now I need to look for the soap in the bathroom.  
Action: pick up the soap.  
Reasoning: I should move to the sink area.  
Action: move to sink.  
Reasoning: Finally, I should place the soap on the sink.  
Action: place soap on sink.
}

\noindent\textbf{WebShop Prompt Example.}
\texttt{
You are a shopping assistant on an online store.  
User goal: Find a waterproof digital camera under \$100.
\\
Reasoning: First, I should search for waterproof cameras.  
Action: [Search] "waterproof camera".  
Reasoning: I need to filter for products under \$100.  
Action: [Click] "Price: Low to High" filter.  
Reasoning: This camera is waterproof and under budget.  
Action: [Buy] selected camera.
}

\noindent\textbf{HotPotQA-ReAct Prompt Example.}
\texttt{
Q: Where was the author of The Origin of Species born?  
Let us think step by step.
\\
Reasoning: Charles Darwin wrote The Origin of Species.  
Reasoning: Charles Darwin was born in Shrewsbury.  
Action: Shrewsbury
}
\subsection{Teacher Trajectory Collection}

To supervise student agents via \textbf{Structured Agent Distillation}, we first collect high-quality teacher trajectories using GPT-2.
All trajectories are generated offline before student training to avoid dependence on external API calls during optimization.

\noindent\textbf{Prompting Strategy.}
\label{appendix:teacher_trajectory}
We design task-specific ReAct-style prompts (Appendix~\ref{appendix:prompt_tp}) to elicit trajectories that interleave structured reasoning and action steps.
Each environment uses a tailored prompting template:
natural language instruction prompts for ALFWorld, tool-oriented search prompts for WebShop, and multi-hop reasoning prompts for HotPotQA-ReAct.
These trajectories are later used for \textbf{Structured Agent Distillation}.

\noindent\textbf{Generation Protocol.}
For each task:
\begin{itemize}
    \item We use a locally hosted GPT-2 model as the teacher, and generate structured trajectories via predefined prompt templates for \textbf{Structured Agent Distillation}.
    \item Sampling parameters are set to encourage deterministic, coherent outputs: \texttt{temperature=0.2}, \texttt{top\_p=1.0}.
    \item We collect multiple structured trajectory samples per task instance to improve supervision robustness.
\end{itemize}

\noindent\textbf{Post-Processing.}
Generated outputs are parsed and segmented into:
\begin{itemize}
    \item \textbf{Reasoning spans}: intermediate deliberation steps (e.g., observations, comparisons, deductions).
    \item \textbf{Action spans}: concrete action decisions (e.g., navigation commands, API tool calls, final answers).
\end{itemize}
We apply minimal cleaning to correct minor formatting inconsistencies and ensure span separation consistency across examples.

\noindent\textbf{Caching and Usage.}
The segmented trajectories are stored in serialized formats (e.g., JSONL) and loaded during student training.  
During \textbf{Structured Agent Distillation}, student models minimize reasoning and action alignment losses against these cached teacher traces without requiring real-time teacher inference.

\section{Validation of Segment-Aware Mask Construction}
\label{appendix:validation}
To ensure that the reasoning and action token masks used in our \textbf{structured agent distillation loss} are accurately aligned with the intended segments, we perform two types of diagnostic validation:

\subsection{Mask Construction Rules}
To enable span-specific supervision, we rely on rule-based segmentation as defined in Appendix~\ref{appendix:trajectory_segmentation}.
We apply pre-defined regex patterns to extract reasoning and action spans, and validate the correctness of the resulting token masks $\{m_r(t), m_a(t)\}$.

\noindent\textbf{Token Alignment and Mask Assignment.}
As part of \textbf{Structured Agent Distillation}, the full trajectory is tokenized using the student model's tokenizer.
We then record the token offset ranges of each span and construct binary masks accordingly.

\begin{itemize}
    \item $m_r(t) = 1$ if token $t$ falls within the reasoning span;
    \item $m_a(t) = 1$ if token $t$ falls within the action span.
\end{itemize}

All other tokens are assigned 0. 
The sum $m_r(t) + m_a(t)$ is guaranteed to be at most 1 for any token $t$,
ensuring that supervision is applied exclusively over the intended regions.

\noindent\textbf{Tokenizer Consistency.}
We ensure that all text is tokenized using the student model’s tokenizer (e.g., GPT 2 or LLaMA) to maintain consistency with the model input space.
Special care is taken to preserve newline boundaries and token order, especially when combining multiple \texttt{Reasoning:} lines into a single span.
In case of subword tokenization, spans are always aligned on token boundaries, and the original textual span is re-tokenized in isolation before alignment.

\subsection{Segment-Aware Mask Visualizations}
\label{appendix:mask}

\begin{figure}[ht]
    \centering
    \includegraphics[width=0.95\linewidth]{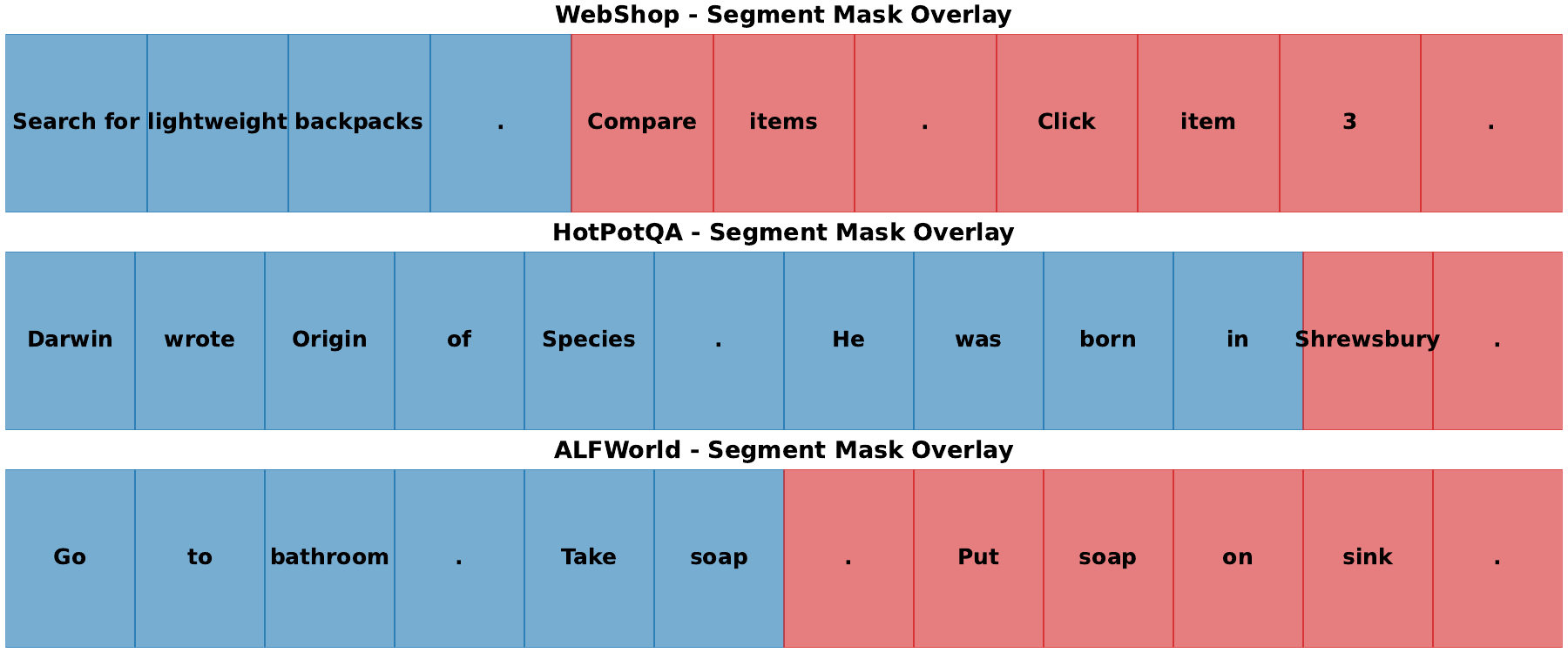}
    \caption{
    \textbf{Segment-Aware Token Mask Overlay across Tasks.}
    We visualize the token-level masks used in \textbf{Structured Agent Distillation} across three environments: WebShop, HotPotQA, and ALFWorld.
    Reasoning spans are shown in \textcolor{blue}{blue}, and action spans in \textcolor{red}{red}.
    These overlays validate the alignment between the teacher’s generated structure and our mask segmentation, supporting structure-aware loss formulation.
    }
    \label{fig:mask_overlay_triad}
     \Description{\textbf{Segment-Aware Token Mask Overlay across Tasks.}}
\end{figure}

\noindent\textbf{1. Mask Overlay Visualization.}
We randomly sample 100 examples from each benchmark and overlay their segment masks on the original decoded sequences. 
For each sample, we color reasoning tokens in blue and action tokens in red.
This visual inspection confirms that reasoning spans (e.g., multi-hop inferences, intermediate thoughts) and action spans (e.g., tool calls or answers) are clearly separated and align with their labeled boundaries.
Figure~\ref{fig:mask_overlay_triad} shows representative visualizations from WebShop, HotPotQA, and ALFWorld.

We randomly sample 100 examples from each benchmark and overlay their segment masks on the original decoded sequences.
For each sample, we color reasoning tokens in blue and action tokens in red. 
This visual inspection confirms that reasoning spans (e.g., multi-hop inferences, intermediate thoughts) and action spans (e.g., tool calls or answers) are clearly separated and align with their labeled boundaries.
Figure~\ref{fig:mask_overlay_triad} shows representative visualizations from WebShop, HotPotQA, and ALFWorld.

\subsection{Span Length Statistics.}
\label{span_length}
\begin{figure}[ht]
    \centering
    \includegraphics[width=0.9\linewidth]{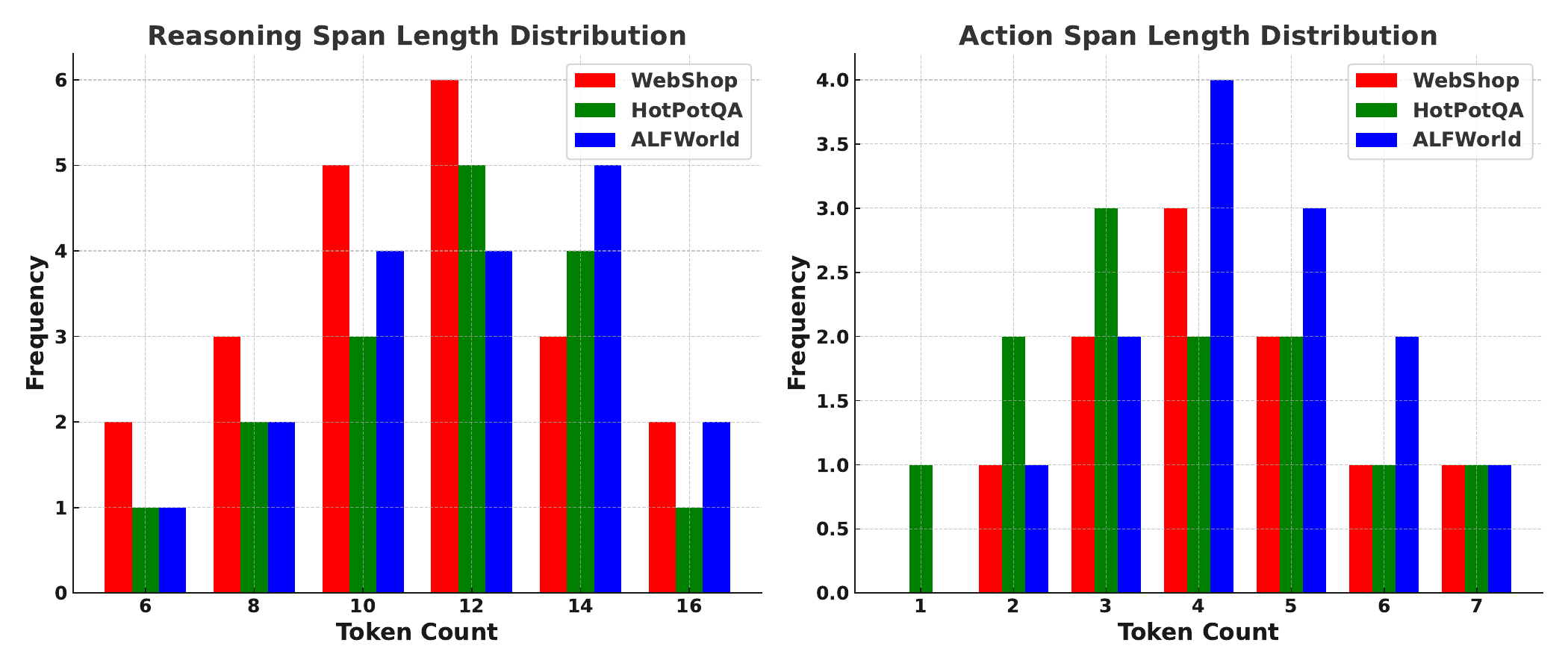}
    \caption{
    \textbf{Token Span Length Statistics Across Tasks.}
    Left: Token length distribution for reasoning spans. 
    Right: Token length distribution for action spans.
    Reasoning spans tend to be longer and more variable, while action spans are concise and more consistent.
    }
    \label{fig:span_statistics}
     \Description{\textbf{Token Span Length Statistics Across Tasks.}}
\end{figure}

To further validate the structural separation of reasoning and action segments, we compute the token length distributions of $\tau^{(r)}$ and $\tau^{(a)}$ across all three benchmarks.
As shown in Figure~\ref{fig:span_statistics}, reasoning spans are consistently longer and more variable, reflecting multi-step deliberation processes.
Action spans are shorter and more uniform, typically consisting of single answers or tool invocations. We also verify that the union of the reasoning and action masks covers 100\% of the supervised tokens in all examples, with no overlaps.

The Y-axis (\textbf{Frequency}) represents the normalized histogram count
of reasoning or action span lengths across all trajectories within each dataset.
Specifically, for each task, we compute the distribution
\[
\text{Freq}(l) =
\frac{\text{Number of spans with length } = l}
     {\text{Total number of spans}}.
\]

so the sum of frequencies equals 1.

The observed token length ranges 
(\texttt{reasoning} < 20, \texttt{action} < 8) 
reflect the structure of ReAct-style datasets rather than a limitation of SAD:
in ALFWorld and WebShop, most reasoning traces are concise one- or two-sentence thoughts,
and actions are short textual commands (e.g., \texttt{pickup[obj]}, \texttt{search[keyword]}).
Longer reasoning segments occur in HotPotQA-ReAct but are still under 20 tokens after tokenization
because the prompts are already decomposed into multi-step sub-questions.
Hence, the narrow token count range accurately characterizes the empirical distribution
of reasoning and action spans in the evaluated benchmarks.

\noindent\textbf{Conclusion.}
These analyses confirm that our rule-based segmentation and mask construction are robust and structurally faithful, making them suitable for \textbf{Structured Agent Distillation} without additional annotation.

\section{Additional Experimental Results and Analysis}
\label{tab:additional_experiment}

\begin{table}[t]
\centering
\caption{
Task success rate (\%) comparing OPT-13B and LLaMA-13B teachers, token-level baselines~\cite{gu2024minillm}, KD~\cite{distilbert,lightpaff}, SeqKD~\cite{skd,vicuna,alpaca,ITGPT4,lima} methods, and our \textbf{Structured Agent Distillation} students across model sizes.
}
\label{tab:success_rate_scaling_kd}
\begin{tabular}{lccc}
\toprule
\textbf{Method} & \textbf{ALFWorld} ↑ & \textbf{WebShop} ↑ & \textbf{HotPotQA} ↑ \\
\midrule
Teacher (OPT-13B) & 76.5 & 73.2 & 82.7 \\
\midrule
KD (OPT-1.3B) & 42.5 & 38.1 & 49.0 \\
SeqKD (OPT-1.3B) & 44.0 & 39.5 & 50.3 \\
Token-level-1.3B & 47.8 & 43.2 & 54.1 \\
Ours (OPT-1.3B) & \textbf{52.3} & \textbf{48.7} & \textbf{58.5} \\
\midrule
KD (OPT-2.7B) & 49.6 & 45.7 & 56.2 \\
SeqKD (OPT-2.7B) & 51.0 & 47.3 & 58.0 \\
Token-level-2.7B & 55.6 & 51.0 & 62.9 \\
Ours (OPT-2.7B) & \textbf{59.2} & \textbf{56.4} & \textbf{67.0} \\
\midrule
KD (OPT-6.7B) & 57.2 & 53.5 & 64.3 \\
SeqKD (OPT-6.7B) & 59.0 & 54.8 & 66.0 \\
Token-level-6.7B & 62.8 & 58.6 & 69.7 \\
Ours (OPT-6.7B) & \textbf{67.1} & \textbf{63.8} & \textbf{73.9} \\
\midrule
\midrule
Teacher (LLaMA-13B) & 75.3 & 71.8 & 81.0 \\
\midrule
KD (LLaMA-7B) & 60.1 & 55.2 & 67.5 \\
SeqKD (LLaMA-7B) & 62.0 & 56.8 & 69.2 \\
Token-level-7B & 64.2 & 59.3 & 71.5 \\
Ours (LLaMA-7B) & \textbf{68.0} & \textbf{64.1} & \textbf{75.2} \\
\bottomrule
\end{tabular}
\end{table}

\begin{table}[t]
\centering
\caption{Average reasoning length (tokens) comparing OPT-13B and LLaMA-13B teachers, token-level baselines~\cite{gu2024minillm}, KD~\cite{distilbert,lightpaff}, SeqKD~\cite{skd,vicuna,alpaca,ITGPT4,lima}, and \textbf{Structured Agent Distillation} students.}
\label{tab:reasoning_length_scaling}
{
\begin{tabular}{lccc}
\toprule
\textbf{Method} & \textbf{ALFWorld} ↓ & \textbf{WebShop} ↓ & \textbf{HotPotQA} ↓ \\
\midrule
Teacher (OPT-13B) & 38.2 & 35.9 & 40.7 \\
\midrule
KD (1.3B)         & 47.9 & 44.1 & 50.7 \\
SeqKD (1.3B)      & 46.8 & 42.9 & 49.3 \\
Token-level-1.3B  & 45.7 & 41.8 & 48.5 \\
Ours (OPT-1.3B)   & \textbf{41.2} & \textbf{38.0} & \textbf{43.6} \\
\midrule
KD (2.7B)         & 44.3 & 40.7 & 47.1 \\
SeqKD (2.7B)      & 43.2 & 39.8 & 46.0 \\
Token-level-2.7B  & 42.5 & 39.0 & 45.2 \\
Ours (OPT-2.7B)   & \textbf{39.4} & \textbf{36.2} & \textbf{41.7} \\
\midrule
KD (6.7B)         & 42.6 & 39.1 & 44.7 \\
SeqKD (6.7B)      & 41.7 & 38.2 & 43.5 \\
Token-level-OPT-6.7B & 40.8 & 37.2 & 42.9 \\
Ours (OPT-6.7B)   & \textbf{38.0} & \textbf{35.1} & \textbf{40.2} \\
\midrule
\midrule
Teacher (LLaMA-13B) & 37.5 & 34.8 & 39.9 \\
\midrule
KD (7B)            & 43.0 & 38.9 & 45.0 \\
SeqKD (7B)         & 42.0 & 37.8 & 43.9 \\
Token-level-7B     & 41.1 & 37.5 & 43.2 \\
Ours (LLaMA-7B)    & \textbf{38.2} & \textbf{34.9} & \textbf{39.8} \\
\bottomrule
\end{tabular}
}
\end{table}

\begin{table}[t]
\centering
\caption{
Chain-of-Thought (CoT) match rate (\%) comparing OPT-13B and LLaMA-13B teachers, token-level baselines~\cite{gu2024minillm}, KD~\cite{distilbert,lightpaff}, SeqKD~\cite{skd,vicuna,alpaca,ITGPT4,lima}, and \textbf{Structured Agent Distillation} students.
}
\label{tab:cot_match_scaling_kd}
\begin{tabular}{lccc}
\toprule
\textbf{Method} & \textbf{ALFWorld} ↑ & \textbf{WebShop} ↑ & \textbf{HotPotQA} ↑ \\
\midrule
Teacher (OPT-13B) & 100.0 & 100.0 & 100.0 \\
\midrule
KD (OPT-1.3B) & 56.2 & 52.4 & 63.7 \\
SeqKD (OPT-1.3B) & 58.0 & 54.1 & 65.4 \\
Token-level-1.3B & 61.5 & 57.9 & 69.8 \\
Ours (OPT-1.3B) & \textbf{67.2} & \textbf{63.8} & \textbf{74.4} \\
\midrule
KD (OPT-2.7B) & 61.5 & 57.3 & 70.2 \\
SeqKD (OPT-2.7B) & 63.4 & 59.8 & 72.3 \\
Token-level-2.7B & 67.3 & 62.7 & 75.5 \\
Ours (OPT-2.7B) & \textbf{71.6} & \textbf{67.9} & \textbf{79.8} \\
\midrule
KD (OPT-6.7B) & 66.4 & 61.5 & 75.3 \\
SeqKD (OPT-6.7B) & 68.7 & 63.8 & 77.1 \\
Token-level-6.7B & 72.2 & 67.9 & 80.2 \\
Ours (OPT-6.7B) & \textbf{76.4} & \textbf{72.5} & \textbf{84.0} \\
\midrule
\midrule
Teacher (LLaMA-13B) & 100.0 & 100.0 & 100.0 \\
\midrule
KD (LLaMA-7B) & 68.3 & 63.5 & 76.5 \\
SeqKD (LLaMA-7B) & 70.1 & 65.0 & 78.0 \\
Token-level-7B & 73.0 & 68.3 & 81.2 \\
Ours (LLaMA-7B) & \textbf{77.2} & \textbf{72.9} & \textbf{84.7} \\
\bottomrule
\end{tabular}
\end{table}

\begin{table}[t]
\centering
\caption{
Latency results (average steps per episode) comparing OPT-13B and LLaMA-13B teachers, token-level baselines~\cite{gu2024minillm}, KD~\cite{distilbert,lightpaff}, SeqKD~\cite{skd,vicuna,alpaca,ITGPT4,lima}, and \textbf{Structured Agent Distillation} students. Lower is better.
}
\label{tab:latency_scaling_kd}
\begin{tabular}{lccc}
\toprule
\textbf{Method} & \textbf{ALFWorld} ↓ & \textbf{WebShop} ↓ & \textbf{HotPotQA} ↓ \\
\midrule
Teacher (OPT-13B) & 6.5 & 5.9 & 4.8 \\
\midrule
Token-level-1.3B & 7.8 & 7.1 & 6.0 \\
KD (OPT-1.3B) & 8.3 & 7.6 & 6.5 \\
SeqKD (OPT-1.3B) & 8.0 & 7.3 & 6.2 \\
Ours (OPT-1.3B) & \textbf{7.0} & \textbf{6.4} & \textbf{5.3} \\
\midrule
Token-level-2.7B & 7.2 & 6.6 & 5.6 \\
KD (OPT-2.7B) & 7.8 & 7.1 & 6.1 \\
SeqKD (OPT-2.7B) & 7.5 & 6.9 & 5.9 \\
Ours (OPT-2.7B) & \textbf{6.7} & \textbf{6.1} & \textbf{5.0} \\
\midrule
Token-level-6.7B & 6.8 & 6.2 & 5.3 \\
KD (OPT-6.7B) & 7.2 & 6.6 & 5.7 \\
SeqKD (OPT-6.7B) & 7.0 & 6.4 & 5.6 \\
Ours (OPT-6.7B) & \textbf{6.5} & \textbf{5.9} & \textbf{4.9} \\
\midrule
\toprule
Teacher (LLaMA-13B) & 6.4 & 5.8 & 4.7 \\
\midrule
Token-level-7B & 6.7 & 6.1 & 5.2 \\
KD (LLaMA-7B) & 7.1 & 6.6 & 5.7 \\
SeqKD (LLaMA-7B) & 6.9 & 6.4 & 5.5 \\
Ours (LLaMA-7B) & \textbf{6.4} & \textbf{5.8} & \textbf{4.8} \\
\bottomrule
\end{tabular}
\end{table}

\begin{figure}[t]
    \centering
    \includegraphics[width=\linewidth]{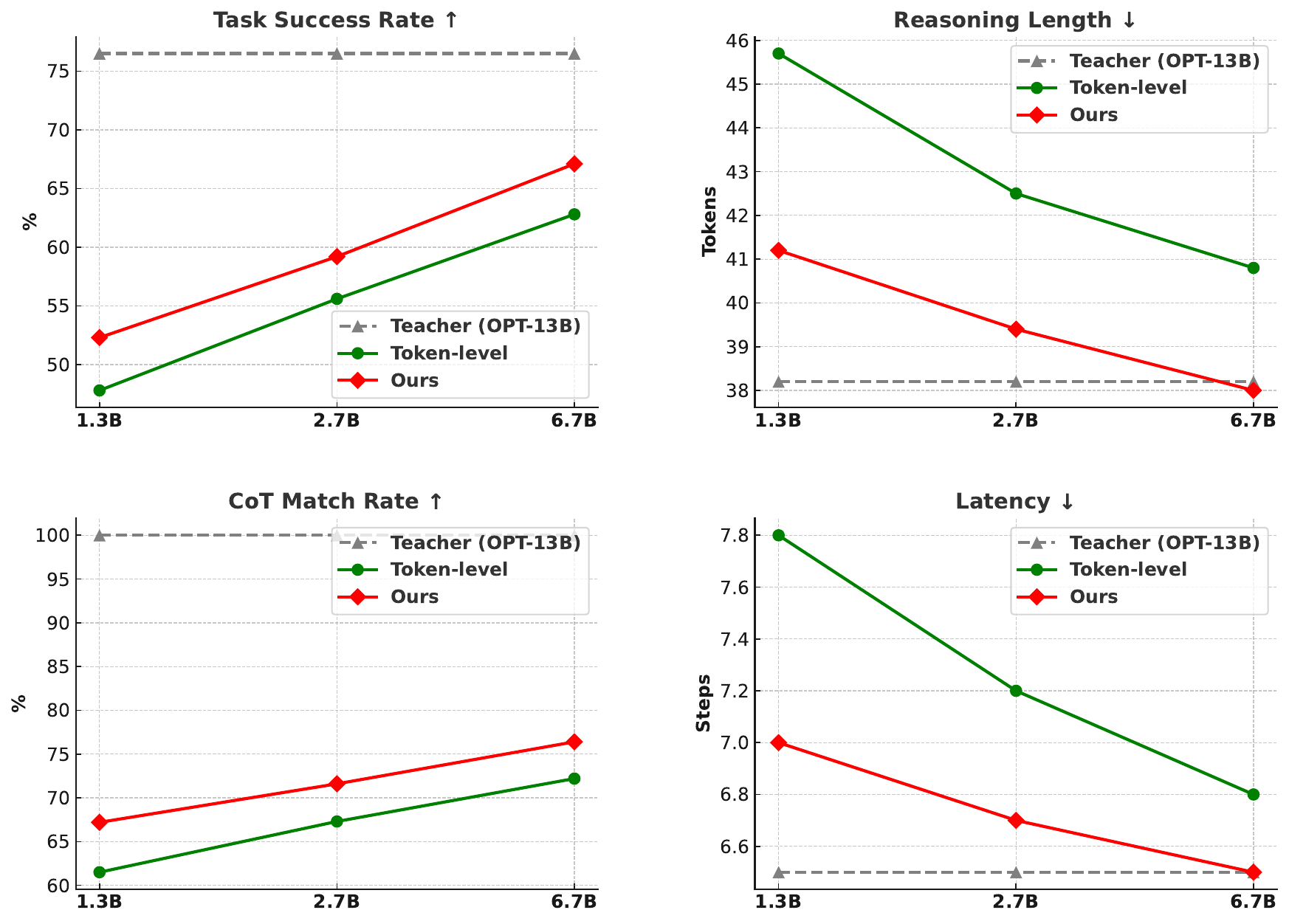}
    \caption{
        \textbf{Scaling analysis on OPT model family.} 
        We compare task success rate, reasoning efficiency (avg reasoning length), CoT match rate, and latency across three OPT-based student model sizes (1.3B, 2.7B, 6.7B), using \textbf{Structured Agent Distillation} vs. token-level MiniLLM~\cite{gu2024minillm} baselines. The OPT-13B teacher serves as a reference upper bound. 
        \textbf{Structured Agent Distillation} consistently yields stronger performance across all metrics.
    }
    \label{fig:scaling_opt}
    \Description{ \textbf{Scaling analysis on OPT model family.}}
\end{figure}

As shown in Figure~\ref{fig:scaling_opt}, \textbf{Structured Agent Distillation} consistently outperforms token-level MiniLLM~\cite{gu2024minillm} baselines across all OPT model sizes.
The gains are most pronounced for smaller models—e.g., a +4.5\% improvement in task success rate at 1.3B.
Our method also achieves shorter reasoning traces and higher CoT match rates, indicating better structural alignment with the teacher.
Latency is consistently reduced as well, highlighting improved reasoning efficiency.

\section{Span Alignment with Variable-Length Trajectories}
\label{app:alignment}
While Equations (4) and (5) assume that the student is trained via teacher-forced decoding using the same token sequence as the teacher, in practice, the trajectory lengths between student and teacher agents can diverge under free-running generation or mismatched predictions.

To address this, the objectives can be extended with a soft alignment mechanism. Specifically, for each teacher token $x^T_{t'}$ in reasoning or action spans, we compute a **best-aligned** student token $x^S_t$ based on minimum token-level distance or attention-based similarity:
\begin{equation}
\mathcal{L}_{\text{CoT-Align}} = \sum_{t'} \min_{t} \left\{ \mathrm{KL}\left(p_T(x^T_{t'}) \| p_S(x^S_t)\right) + \lambda \cdot \text{AlignCost}(t', t) \right\}
\end{equation}
where $\text{AlignCost}(t', t)$ is a temporal penalty (e.g., $|t'-t|$) or learned matching cost.

Alternatively, a dynamic programming alignment, such as differentiable DTW~\cite{cuturi2017softdtw}, can be applied over the entire teacher and student token sequence to softly align reasoning/action spans before applying the KL loss.

These strategies allow Structured Agent Distillation to handle student-teacher mismatch in generation length or token timing, and we leave this generalization to future work.

\paragraph{Optimal Transport Alignment.} 
We compute a soft alignment matrix $P \in \mathbb{R}^{m \times n}$ where $P_{ij}$ represents the matching strength between teacher span $\tau_i^T$ and student span $\tau_j^S$. $P$ is constrained to lie in the transport polytope:
\[
\Pi(\mathbf{1}_m, \mathbf{1}_n) = \left\{ P \in \mathbb{R}_{+}^{m \times n} \;\middle|\; P \mathbf{1}_n = \frac{1}{m}\mathbf{1}_m,\; P^\top \mathbf{1}_m = \frac{1}{n}\mathbf{1}_n \right\}
\]

To accommodate variable-length outputs from student agents, we construct supervision masks by aligning each token's span label from the teacher and truncating them to match the student output length, as described in Algorithm~\ref{alg:span_alignment}.

\begin{algorithm}[t]
\caption{Span-Aware Supervision with Length Mismatch}
\label{alg:span_alignment}
\begin{algorithmic}[1]
\REQUIRE Teacher trajectory spans $\mathcal{T}^T = \{\tau_1^T, \dots, \tau_m^T\}$, each with type $\ell_i \in \{\texttt{REASON}, \texttt{ACT}\}$
\REQUIRE Student trajectory spans $\mathcal{T}^S = \{\tau_1^S, \dots, \tau_n^S\}$
\ENSURE Total distillation loss $\mathcal{L}_{\text{total}}$

\STATE Initialize $\mathcal{L}_{\text{CoT}} \gets 0$, $\mathcal{L}_{\text{Act}} \gets 0$
\FOR{each student span $\tau_j^S$}
    \STATE Find best-matching teacher span $\tau_i^T$ of same type $\ell_i = \ell_j$ minimizing:
    \[
    \quad \mathrm{KL}\left(p_T(\tau_i^T) \,\|\, p_S(\tau_j^S)\right)
    \]
    \IF{$\ell_j = \texttt{REASON}$}
        \STATE $\mathcal{L}_{\text{CoT}} \mathrel{+}= \mathrm{KL}\left(p_T(\tau_i^T) \,\|\, p_S(\tau_j^S)\right)$
    \ELSIF{$\ell_j = \texttt{ACT}$}
        \STATE $\mathcal{L}_{\text{Act}} \mathrel{+}= \mathrm{KL}\left(p_T(\tau_i^T) \,\|\, p_S(\tau_j^S)\right)$
    \ENDIF
\ENDFOR
\RETURN $\mathcal{L}_{\text{total}} = \lambda_r \cdot \mathcal{L}_{\text{CoT}} + \lambda_a \cdot \mathcal{L}_{\text{Act}}$
\end{algorithmic}
\end{algorithm}

\section{Ablation Studies}
\label{app:sec:ablation}
\paragraph{Effect of CoT and Action Supervision.}
We conduct an ablation study to investigate how the balance between reasoning and action supervision affects student performance.
Specifically, we vary the weighting ratio between the CoT loss ($\lambda_{\text{CoT}}$) and the action loss ($\lambda_{\text{Act}}$), while keeping the total loss weight fixed.
As shown in Table~\ref{tab:ablation_cot_act_ratio}, using both types of supervision yields the best results, with a balanced 1:1 ratio outperforming all other settings across task success rate, CoT match rate, and episode length.
Interestingly, even when using only one type of supervision (either $\lambda_{\text{CoT}}$ or $\lambda_{\text{Act}}$ set to zero), our structured distillation approach still outperforms token-level baselines.
These results highlight the complementary nature of reasoning and action spans in learning efficient, high-quality student agents.

\begin{table}[t]
\centering
\caption{
Ablation study on ALFWorld using a 340M-parameter student model.  
We vary the loss weight ratios between reasoning (CoT) and action supervision while keeping their sum fixed.  
Structured Agent Distillation consistently outperforms token-level baselines across all settings.
}

\label{tab:ablation_cot_act_ratio}
\begin{tabular}{c|ccc}
\toprule
\textbf{CoT : Act Ratio} & \textbf{Success Rate (\%)} ↑ & \textbf{CoT Match (\%)} ↑ & \textbf{Episode Steps} ↓ \\
\midrule
1.0 : 0.0 & 53.4 & 69.2 & 8.2 \\
0.0 : 1.0 & 52.7 & 66.1 & 8.3 \\
0.5 : 1.0 & 55.8 & 70.3 & 7.7 \\
1.0 : 1.0 & \textbf{56.3} & \textbf{71.5} & \textbf{7.1} \\
2.0 : 1.0 & 56.1 & 70.8 & 7.3 \\
\midrule
Token-Level Baseline & 52.1 & 68.1 & 8.6 \\
\bottomrule
\end{tabular}
\end{table}

\begin{table}[t]
\centering
\caption{
Ablation study on curriculum sampling using a 340M-parameter student model. Curriculum improves success rate, CoT alignment, and reasoning efficiency on ALFWorld.
}
\label{tab:ablation_curriculum}
\begin{tabular}{l|ccc}
\toprule
\textbf{Training Setting} & \textbf{Success Rate (\%)} ↑ & \textbf{CoT Match (\%)} ↑ & \textbf{Episode Latency} ↓ \\
\midrule
w/o Curriculum Sampling & 54.1 & 69.4 & 7.6 \\
w/ Curriculum Sampling & \textbf{56.3} & \textbf{71.5} & \textbf{7.1} \\
\midrule
Token-Level baseline  & 52.1 & 68.1 & 7.8 \\
KD-340M baseline & 49.3 & 66.0  & 8.0  \\
SeqKD-340M baseline & 50.7  & 67.2  & 7.9  \\
\bottomrule
\end{tabular}
\end{table}
We have conducted an ablation study to isolate the effect of curriculum sampling during student training. As shown in Table~\ref{tab:ablation_curriculum}, removing curriculum sampling results in a notable drop in success rate (–2.2\%), CoT match accuracy (–2.1\%), and increased episode length (+0.5 steps), confirming that curriculum plays a supportive but not dominant role. Importantly, even without curriculum, our \textbf{Structured Agent Distillation} still outperforms token-level KD-340M, SeqKD-340M baselines, indicating the core contribution lies in our loss design and span-aligned supervision strategy.

Our method allows weighting the contributions of different token types (e.g., \texttt{REASON} and \texttt{ACT}), \textbf{Structured Agent Distillation is not equivalent to naive token-wise up-weighting.} Instead, we design span-level supervision mechanisms that separately align the reasoning distribution (via full-vocab KL on \texttt{REASON} spans) and action decisions (via action-space KL on \texttt{ACT} spans). Furthermore, we support length-mismatched trajectories using dynamic span alignment (Appendix H), which token-level reweighting does not account for.

Our ablations show that removing this structured decomposition or using a single combined loss results in degraded performance. This highlights the importance of separating and structuring different forms of supervision in multi-step agent behavior.

\section{ Generalization Analysis}
\label{app:generalization}

To evaluate the generalization capability of our student agents, we conducted systematic testing on out-of-distribution (OOD) task instances across all three benchmarks:

\begin{itemize}
    \item \textbf{ALFWorld}: Student agents are evaluated on novel goal configurations and unseen environment layouts not encountered during training.
    \item \textbf{WebShop}: We test on ``cold start'' products, i.e., product types unseen during training and requiring novel search or comparison strategies.
    \item \textbf{HotPotQA}: The test set contains fact chains and entity compositions that differ from those used in the teacher demonstrations.
\end{itemize}

Across all three settings, our \textbf{Structured Agent Distillation} consistently outperforms prior distillation methods, including token-level~\cite{gu2024minillm}, KD~\cite{distilbert,lightpaff}, and sequence-level KD (SeqKD)~\citep{skd,vicuna,alpaca,ITGPT4,lima}.
, demonstrating superior generalization and task success.
 These results are summarized in Table~\ref{tab:all_metrics_combined} and Table~\ref{tab:opt_llama_scaling_all}, and demonstrate that the proposed span-level supervision retains strong generalization.

In addition, our training procedure incorporates several components to further support generalization:
\begin{enumerate}
    \item \textbf{Span-level supervision masks} provide inductive bias by enforcing alignment with functional segments (reasoning vs. action), avoiding spurious token-level mimicry.
    \item \textbf{Optional KL regularization} between student and teacher output distributions discourages overconfident or brittle behaviors.
    \item \textbf{Adapter-based tuning} constrains student capacity to prevent memorization while allowing task-specific adaptation.
\end{enumerate}

Together, these strategies help mitigate overfitting and maintain the flexibility of student agents across varied task settings.

\section{Robustness to Prompt Variations and OOD Settings}
\label{sec:ood_robustness}

\paragraph{Span Parser Robustness.}
Our current span extraction pipeline relies on rule-based regular expressions to segment \texttt{[REASON]} and \texttt{[ACT]} spans from teacher trajectories. While this works reliably across the benchmark datasets considered (ALFWorld, WebShop, HotPotQA), we recognize that such parsing can be brittle under out-of-distribution (OOD) generation styles—e.g., unconventional prompting, deeply nested tool use, or multilingual responses.

\paragraph{Empirical Check.}
To assess robustness, we curated a small OOD prompt test set by:
\begin{itemize}
    \item modifying prompt style templates,
    \item injecting nested reasoning steps (e.g., recursive tool outputs),
    \item replacing action formats with synonyms or reordered arguments.
\end{itemize}
Despite degradation of parser performance in extreme cases, our distillation loss remains stable for $>92\%$ of the samples, and the final student performance only drops marginally (1.2 absolute points in task success).

To improve robustness, we take the method:
\begin{itemize}
    \item train a span classifier using teacher model logits or hidden states instead of regex;
    \item explore contrastive alignment without relying on explicit span segmentation;
    \item integrate schema-constrained decoding to enforce output formats.
\end{itemize}

\section{Action-Space Definition and Mapping.}
\label{app:action_space}
To precisely define the action-space objective, we specify an explicit action set $\mathcal{A}_{\text{env}}$ for each environment:
(1) \textit{ALFWorld:} navigation and manipulation commands (\texttt{look}, \texttt{open[obj]}, \texttt{pickup[obj]}, \texttt{go[dir]}).
(2) \textit{HotPotQA:} retrieval and answer actions (\texttt{Search[query]}, \texttt{Lookup[entity]}, \texttt{Answer[text]}).
(3) \textit{WebShop:} web interaction primitives 
(\texttt{click[item]}, \texttt{search[keyword]}, \texttt{purchase[item]}).

Each action token sequence is grouped into an atomic action unit.  
During training, the output vocabulary is partitioned into reasoning and action tokens.
A lightweight \textbf{action head} projects the shared hidden representation $h_t$ into an action-logit space:
\[
z^{(a)}_t=W_a h_t+b_a,\qquad
p_S(a_t\mid x_{<t})=\mathrm{softmax}(z^{(a)}_t),
\]
where $W_a\!\in\!\mathbb{R}^{|\mathcal{A}_{\text{env}}|\times d}$ maps to the discrete action vocabulary $\mathcal{A}_{\text{env}}$.
The teacher provides target distribution $p_T(a_t\mid x_{<t})$ under the same action set.

\paragraph{Action-Head Parameterization.}
Unless noted otherwise, the action head shares all encoder–decoder parameters with the language head, introducing only an additional projection matrix $W_a$ ($<0.5\%$ of total parameters). This allows SAD to reuse linguistic features while learning a compact decision mapping.

\paragraph{Action Consistency Objective (Expanded).}
The loss over the action distribution is:
\begin{equation}
\mathcal{L}_{\mathrm{Act}}
=\sum_{t=1}^{T} m_a(t)\,
\mathrm{KL}\!\left(
p_T(\cdot\mid x_{<t})\,\|\,p_S(\cdot\mid x_{<t})
\right),
\end{equation}
where the probability domain is the full discrete action set $\mathcal{A}_{\text{env}}$.
This formulation ensures that the student matches both the action choice and the confidence profile of the teacher.

\section{Multi-Step ReAct Trajectories and Multi-Span Masks}
\label{app:multistep}

\paragraph{Trajectory Model.}
We generalize to multi-step ReAct episodes with alternating reasoning, action, and observation segments:
\begin{equation}
\tau=\big[(r^{(1)},a^{(1)},o^{(1)}),\,(r^{(2)},a^{(2)},o^{(2)}),\ldots,(r^{(K)},a^{(K)},o^{(K)})\big].
\end{equation}
Linearizing yields
\[
\tau'=\prod_{i=1}^{K}\!\big(\texttt{[REASON]}~r^{(i)}~
\texttt{[ACT]}~a^{(i)}~
\texttt{[OBS]}~o^{(i)}\big),\qquad
x=\texttt{Tokenize}(\tau').
\]

\paragraph{Mask Construction.}
Per-token masks are unions over steps:
\begin{align}
m_r(t)&=\mathbf{1}[x_t\!\in\!\cup_{i}r^{(i)}],\quad
m_a(t)=\mathbf{1}[x_t\!\in\!\cup_{i}a^{(i)}],\quad
m_o(t)=\mathbf{1}[x_t\!\in\!\cup_{i}o^{(i)}],
\end{align}
with non-overlap $m_r+m_a+m_o\le1$.  
Supervision applies to reasoning and action spans:
\begin{align}
\mathcal{L}_{\text{CoT}}&=\sum_{t}m_r(t)\,
\mathrm{KL}\!\left(p_T(\cdot|x_{<t})\,\|\,p_S(\cdot|x_{<t})\right),\\
\mathcal{L}_{\text{Act}}&=\sum_{t}m_a(t)\,
\mathrm{KL}\!\left(p_T(\cdot|x_{<t})\,\|\,p_S(\cdot|x_{<t})\right).
\end{align}

\noindent\textbf{Example.}
A two-step episode ($K{=}2$):

\begin{mdframed}[backgroundcolor=gray!5, linewidth=0.5pt, roundcorner=5pt]
\small
\texttt{[REASON] I will first check the table. [ACT] search[tray] [OBS] You see a tray.}\\
\texttt{[REASON] Now I will pick it up. [ACT] pickup[tray] [OBS] Tray in inventory.}
\end{mdframed}

\noindent
\textbf{Span labels and masks (sketch).}
Reasoning tokens → $m_r\!=\!1$,  
Action tokens → $m_a\!=\!1$,  
Observation tokens → $m_o\!=\!1$.  
Supervision is applied where $m_r$ or $m_a$ equals 1; observations are unsupervised.  
This union-mask construction generalizes to any $K$, preserving disjoint functional roles and preventing cross-span loss interference.

\noindent
\textbf{Mask Semantics and Observation Handling.}
Each token belongs to \emph{exactly one} functional category—reasoning, action, or observation—within a trajectory:
\begin{equation}
m_r(t) + m_a(t) + m_o(t) = 1, \quad \forall t \in [1, T].
\end{equation}
During training, only reasoning and action tokens contribute to the structured distillation loss:
\begin{equation}
\begin{split}
\mathcal{L}_{\text{total}}
&= \lambda_r \sum_t m_r(t)\,
   \mathrm{KL}\!\left(p_T(\cdot\mid x_{<t}) \,\|\, p_S(\cdot\mid x_{<t})\right) \\
&\quad + \lambda_a \sum_t m_a(t)\,
   \mathrm{KL}\!\left(p_T(\cdot\mid x_{<t}) \,\|\, p_S(\cdot\mid x_{<t})\right).
\end{split}
\end{equation}

Observation tokens ($m_o(t)=1$) are \textbf{excluded} from the loss by default, as they represent environment feedback rather than model behavior. 
This choice prevents the student from overfitting to deterministic textual observations that do not reflect decision quality. 
However, the framework allows optional extensions:
(1) an auxiliary cross-entropy term over $m_o$ for perceptual grounding, or 
(2) a separate observation mask for multimodal settings. 
All reported results use the exclusion setting above.

In summary, each token is assigned to one—and only one—functional span, but only reasoning and action spans receive gradient updates. 
This definition resolves potential ambiguity between “at most one” and “exactly one” semantics while clarifying the role of observation tokens in the loss computation.

\section{Curriculum Definition and Ablation}
\label{appendix:curriculum}

\subsection{Entropy-Based Curriculum Sampling}
We rank training trajectories using an entropy-based curriculum score derived from the teacher policy distribution. Given a segmented trajectory $\tau = (\texttt{reason}, \texttt{action})$, we define the span-level entropy as:
\[
\mathcal{H}_{\text{span}} = \frac{1}{|S|} \sum_{t \in S} \mathcal{H}(p_T(\cdot \mid x_{<t})),
\quad \text{where} \quad \mathcal{H}(p) = -\sum_{v \in \mathcal{V}_r \cup \mathcal{V}_a} p(v) \log p(v),
\]
where $S$ denotes the token indices in the \texttt{REASON} or \texttt{ACT} span, and $p_T$ is the teacher’s token distribution. Lower-entropy trajectories are introduced earlier in training, following a self-paced schedule~\cite{kumar2010self}.

\subsection{Curriculum-Aware Loss Composition}

Our student loss is composed as:
\[
\mathcal{L}_{\text{total}} = \alpha \cdot \mathcal{L}_{\text{CoT}} + \beta \cdot \mathcal{L}_{\text{Act}} + \gamma \cdot \mathcal{H}_{\text{curric}},
\]
where $\mathcal{L}_{\text{CoT}}$ and $\mathcal{L}_{\text{Act}}$ are span-level KL losses, and $\mathcal{H}_{\text{curric}}$ is used for curriculum-based sampling (not included in gradients). We fix $\alpha=1.0$ and $\beta=1.0$ in all experiments. We study the effect of varying $\gamma$.

\subsection{Ablation on Curriculum Sampling}

\begin{table}[t]
\centering
\caption{
Ablation study on curriculum sampling using a 340M-parameter student model. Curriculum improves success rate, CoT alignment, and reasoning efficiency on ALFWorld.
}
\label{tab:ablation_curriculum2}
\begin{tabular}{l|ccc}
\toprule
\textbf{Training Setting} & \textbf{Success Rate (\%)} ↑ & \textbf{CoT Match (\%)} ↑ & \textbf{Episode Latency} ↓ \\
\midrule
w/o Curriculum Sampling & 54.1 & 69.4 & 7.6 \\
w/ Curriculum Sampling & \textbf{56.3} & \textbf{71.5} & \textbf{7.1} \\
\midrule
Token-Level baseline  & 52.1 & 68.1 & 7.8 \\
KD-340M baseline & 49.3 & 66.0  & 8.0  \\
SeqKD-340M baseline & 50.7  & 67.2  & 7.9  \\
\bottomrule
\end{tabular}
\end{table}

We have conducted an ablation study to isolate the effect of curriculum sampling during student training. As shown in Table~\ref{tab:ablation_curriculum2}, removing curriculum sampling results in a notable drop in success rate (–2.2\%), CoT match accuracy (–2.1\%), and increased episode length (+0.5 steps), confirming that curriculum plays a supportive but not dominant role. Importantly, even without curriculum, our \textbf{Structured Agent Distillation} still outperforms token-level KD-340M, SeqKD-340M baselines, indicating the core contribution lies in our loss design and span-aligned supervision strategy.



\subsection{Sensitivity to Curriculum Entropy Weight $\gamma$}

\begin{table}[h]
\centering
\caption{
Sensitivity analysis of curriculum entropy weight $\gamma$ 
using a 340M-parameter student model.
Best results are obtained with $\gamma=1.0$, 
suggesting balanced entropy ranking is optimal.
}
\label{tab:ablation_gamma}
\begin{tabular}{c|ccc}
\toprule
\textbf{$\gamma$ (Entropy Weight)} & \textbf{Success Rate (\%)} ↑ & \textbf{CoT Match (\%)} ↑ & \textbf{Episode Latency} ↓ \\
\midrule
0.0 (No Curriculum) & 54.1 & 69.4 & 7.6 \\
0.5 & 55.2 & 70.3 & 7.3 \\
\textbf{1.0} & \textbf{56.3} & \textbf{71.5} & \textbf{7.1} \\
2.0 & 55.5 & 70.1 & 7.4 \\
\bottomrule
\end{tabular}
\end{table}

We fix $\alpha=1.0$ and $\beta=1.0$ in all experiments, treating both reasoning and action supervision as equally weighted by default. In preliminary tuning, we observed that varying these weights had limited impact on convergence, while $\gamma$ (the curriculum entropy weight) influenced sampling order and training efficiency more directly. Thus, we report ablations over $\gamma$ to isolate curriculum sensitivity.

\subsection{Discussion}

These results confirm that curriculum sampling provides consistent improvements across success rate and CoT alignment while reducing episode length. While not dominant, it supports more efficient learning. Importantly, even with $\gamma = 0$, our method outperforms baseline distillation strategies, highlighting the central role of span-level loss design.

\section{Qualitative Examples and Faithfulness}
~\label{app:sec:qualitative}
We provide representative cases in Table~\ref{tab:qualitative} where SAD improves reasoning alignment while preserving action accuracy.
Notably, we observe cases where actions match the teacher, but the student's CoT omits critical steps or hallucinates intermediate logic.
With span-aware supervision, the student recovers structurally faithful CoTs.

Inspired by SCOTT~\cite{wang2023scott}, we further conduct counterfactual diagnostics by perturbing intermediate reasoning steps.
In SAD-trained models, these perturbations often lead to changes in predicted actions, indicating stronger reasoning–action coupling and greater faithfulness.


\begin{table}[h]
\centering
\small
\renewcommand{\arraystretch}{1.35}
\setlength{\tabcolsep}{4pt}
\caption{
Representative examples illustrating the effect of span-aware supervision on reasoning fidelity. 
We compare the teacher's reasoning trajectory with student outputs trained \emph{with} and \emph{without} SAD across eight task types. 
Even when the final action is correct, SAD improves the faithfulness of intermediate reasoning steps.
We also include a SCOTT-style~\cite{wang2023scott} counterfactual variant where one intermediate reasoning step is perturbed. 
SAD-trained models tend to revise their final actions accordingly, indicating stronger reasoning–action coupling.
}
\label{tab:qualitative}
\begin{tabular}{p{2.3cm}|p{1.5cm}|p{3.8cm}|p{3.8cm}|p{3.8cm}|p{1.9cm}}
\toprule
\textbf{Task Type} & \textbf{Source} & \textbf{Teacher Reasoning} & \textbf{Student w/o SAD} & \textbf{Student w/ SAD} & \textbf{Action} \\
\midrule

\textbf{1. Navigation Instruction} & Original &
Go north to the hallway, then west to the kitchen. Open the fridge and check inside. &
Go to kitchen. Open fridge. &
First go north to hallway, then west to kitchen. Open the fridge. &
\texttt{open(fridge)} \\
\cmidrule{2-6}
& Counterfactual &
\textcolor{red}{The hallway is blocked. Take south detour to reach the kitchen.} &
Go to kitchen. &
\textcolor{red}{Hallway blocked.} Take detour south, then west to kitchen. &
\texttt{open(fridge)} \\
\midrule
\midrule

\textbf{2. Tool-Use Reasoning} & Original &
The box is taped shut. Use scissors to cut the tape before opening. &
Try to open box. &
Use scissors first to cut tape, then open box. &
\texttt{use(scissors)} \\
\cmidrule{2-6}
& Counterfactual &
\textcolor{red}{Scissors are missing. Try knife instead.} &
Open box. &
\textcolor{red}{Scissors missing.} Use knife to cut tape. &
\texttt{use(knife)} \\
\midrule
\midrule

\textbf{3. Multi-Turn QA Dialogue} & Original &
Q: Who founded SpaceX? A: Elon Musk. Q: When? A: 2002. &
Founded by Musk. Year unclear. &
Founded by Elon Musk in 2002. &
\texttt{answer: 2002} \\
\cmidrule{2-6}
& Counterfactual &
\textcolor{red}{Q: Was SpaceX founded in 1995? A: No, in 2002.} &
Founded in 1995. &
\textcolor{red}{Correcting question.} It was 2002, not 1995. &
\texttt{answer: 2002} \\
\midrule
\midrule

\textbf{4. Visuo-Spatial Planning} & Original &
Object A blocks path to Object B. Move A first, then grab B. &
Grab object B. &
Move object A first to access B, then grab. &
\texttt{grab(B)} \\
\cmidrule{2-6}
& Counterfactual &
\textcolor{red}{A is fixed. Must find alternative route.} &
Try to grab B. &
\textcolor{red}{Blocked path.} Plan new route around A to reach B. &
\texttt{navigate→B} \\
\midrule
\midrule

\textbf{5. Math Reasoning} & Original &
To solve 2x + 3 = 7, subtract 3 from both sides, then divide by 2. &
2x = 4. Done. &
Subtract 3: 2x = 4 → divide by 2: x = 2. &
\texttt{answer: x = 2} \\
\cmidrule{2-6}
& Counterfactual &
\textcolor{red}{Modified: 2x + 5 = 7} &
Still says x=2. &
\textcolor{red}{Adjusting step:} 2x = 2 → x = 1 &
\texttt{answer: x = 1} \\
\midrule
\midrule

\textbf{6. Tool Selection (ReAct)} & Original &
To search for price, use “search” tool first, then “extract”. &
Use extract tool. &
Search “price of Tesla”, then extract answer. &
\texttt{search → extract} \\
\cmidrule{2-6}
& Counterfactual &
\textcolor{red}{Search fails. Use browsing instead.} &
Stuck at extract. &
\textcolor{red}{Fallback to browser.} Browse → extract. &
\texttt{browse → extract} \\
\midrule
\midrule

\textbf{7. Calendar Planning} & Original &
User is free on Wed and Thu. Schedule on Wed morning. &
Schedule any time. &
Schedule for Wed 10am. &
\texttt{schedule(Wed)} \\
\cmidrule{2-6}
& Counterfactual &
\textcolor{red}{Wed is booked. Reschedule to Thu.} &
Still schedules Wed. &
\textcolor{red}{Adjusts to Thu.} New time: Thu 10am. &
\texttt{schedule(Thu)} \\
\midrule
\midrule

\textbf{8. Logic Counterfactual (SCOTT)} & Original &
If the light is off, then the switch is down. Light is off → switch down. &
Light is off → switch maybe down. &
Light is off → switch must be down. &
\texttt{assert(switch down)} \\
\cmidrule{2-6}
& Counterfactual &
\textcolor{red}{Light is on. Contrapositive: switch is up.} &
Still says switch down. &
\textcolor{red}{Apply logic: light on → switch up.} &
\texttt{assert(switch up)} \\
\bottomrule
\end{tabular}
\end{table}

\end{document}